%% file: main.tex
\documentclass[10pt,twocolumn,letterpaper]{article}
\pdfoutput=1

\usepackage{cvpr}            

\usepackage{url}
\usepackage{booktabs}
\usepackage{multicol}
\usepackage{multirow}
\usepackage{makecell}
\usepackage{diagbox}
\usepackage{wrapfig}
\usepackage{tabularx}
\usepackage{array}
\usepackage{graphicx}
\usepackage{marvosym}
\usepackage{hyperref}
\usepackage{xcolor}

\newcommand\Tstrut{\rule{0pt}{2.6ex}}  
\newcommand\Bstrut{\rule[-1.2ex]{0pt}{0pt}}


\definecolor{cvprblue}{rgb}{0.21,0.49,0.74}
\usepackage{hyperref}

\title{VFXMaster: Unlocking Dynamic Visual Effect Generation via\\ In-Context Learning}

\author{
Baolu Li\textsuperscript{1}$^*$, \quad
Yiming Zhang\textsuperscript{1}$^*$, \quad
Qinghe Wang\textsuperscript{1,2}$^*$$^\dagger$, \quad
Liqian Ma\textsuperscript{3}$^\text{\Letter}$, \quad
Xiaoyu Shi\textsuperscript{2}, \quad \\
Xintao Wang\textsuperscript{2}, \quad  
Pengfei Wan\textsuperscript{2}, \quad 
Zhenfei Yin\textsuperscript{4}, \quad
Yunzhi Zhuge\textsuperscript{1}, \quad \\
Huchuan Lu\textsuperscript{1}, \quad
Xu Jia\textsuperscript{1}$^\text{\Letter}$\\
\normalsize$^{1}$Dalian University of Technology ~~
\normalsize$^{2}$Kling Team, Kuaishou Technology~~
\normalsize$^{3}$ZMO AI Inc.~~
\normalsize$^{4}$Oxford University \\[4pt]
\normalsize \href{https://libaolu312.github.io/VFXMaster}{\textcolor{magenta}{https://libaolu312.github.io/VFXMaster}}
}

\begin{document}

\twocolumn[
    {
        \renewcommand\twocolumn[1][]{#1}
        \maketitle
        \centering
        \vspace{-8.0mm}
        \includegraphics[width=0.95\textwidth]{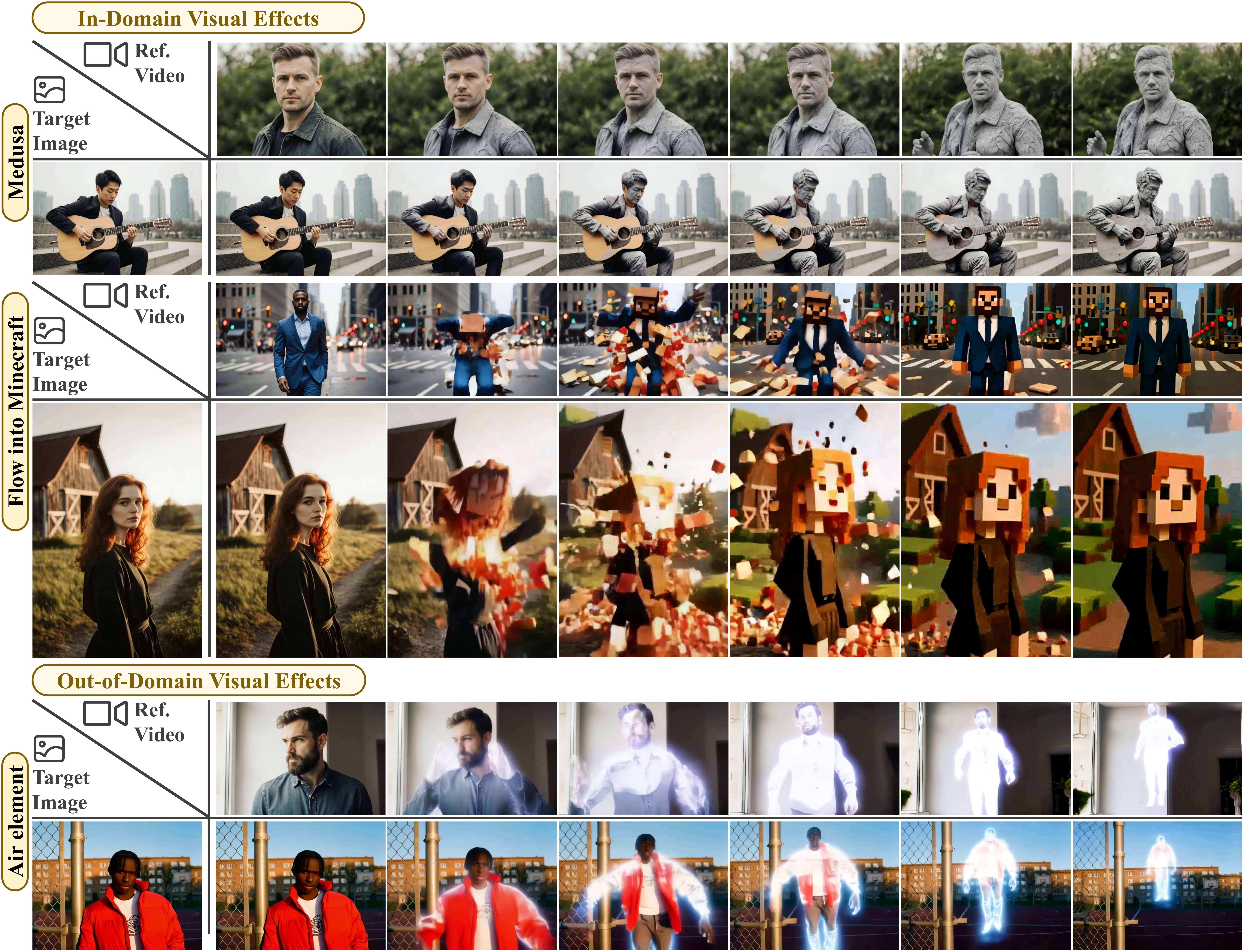}
        \captionsetup{type=figure}
        \vspace{-2.3mm}
        \caption{\textbf{VFXMaster} is a unified reference-based cinematic visual effect~(VFX) generation framework that can reproduce the intricate dynamics and transformations from a reference video onto a user-provided image. It not only shows outstanding performance on in-domain effects, but also strong generalization capability on out-of-domain effects.}
        \label{fig:teaser}
        \vspace{3.0mm}
    }
]

\let\thefootnote\relax\footnotetext{
    $^*$ Equal Contribution\hspace{3pt} 
    \hspace{3pt}$^\dagger$ Project Leader
    \hspace{5pt}$^\text{\Letter}$ Corresponding Author\hspace{3pt}
}

\input{0_abstract}    
\input{1_intro}
\input{2_related}
\input{3_method}
\input{4_exp}
\input{5_conclu}

\input{main.bbl}
\input{X_suppl}

\end{document}

%% file: 0_abstract.tex
\begin{abstract}
Visual effects (VFX) are crucial to the expressive power of digital media, yet their creation remains a major challenge for generative AI. Prevailing methods often rely on the one-LoRA-per-effect paradigm, which is resource-intensive and fundamentally incapable of generalizing to unseen effects, thus limiting scalability and creation. To address this challenge, we introduce VFXMaster, the first unified, reference-based framework for VFX video generation. It recasts effect generation as an in-context learning task, enabling it to reproduce diverse dynamic effects from a reference video onto target content. In addition, it demonstrates remarkable generalization to unseen effect categories. Specifically, we design an in-context conditioning strategy that prompts the model with a reference example. An in-context attention mask is designed to precisely decouple and inject the essential effect attributes, allowing a single unified model to master the effect imitation without information leakage. In addition, we propose an efficient one-shot effect adaptation mechanism to boost generalization capability on tough unseen effects from a single user-provided video rapidly. Extensive experiments demonstrate that our method effectively imitates various categories of effect information and exhibits outstanding generalization to out-of-domain effects. To foster future research, we will release our code, models, and a comprehensive dataset to the community.
\end{abstract}
\vspace{-12pt}

%% file: 1_intro.tex
\section{Introduction}
\label{sec:intro}

Visual effects (VFX) are an integral component of modern digital media, greatly enriching the visual expressiveness of films, games, and social media content. Traditional VFX production is a time-consuming and labor-intensive process that demands specialized skills across multiple stages, including modeling, rigging, animation, rendering, and compositing~\cite{du2021diffpd}. Recent and rapid advancements in generative AI bring revolutionary opportunities for content creation~\cite{ma2025controllable,wang2024stableidentity}. In particular, the growing maturity of video generation models~\cite{yang2024cogvideox,kong2024hunyuanvideo,wan2025wan,HaCohen2024LTXVideo} is ushering in a new era of controllable content synthesis. However, due to data scarcity and the complexity of transformations, the dynamic visual effect generation task has been rarely studied so far.

Existing video generation models, pretrained on large-scale real-world datasets, possess powerful content generation capability. However, VFX often contain anti-physical, surreal, and counter-intuitive elements, such as the particle dynamics of an energy beam or the brilliant patterns of magical elements~\cite{bai2025impossible}. These highly abstract and imaginative concepts represent an out-of-domain challenge that falls significantly outside the knowledge scope of pretrained models. Even with highly detailed text prompts, these models struggle to produce the desired effects accurately. Furthermore, prevailing controllable generation methods focus on spatial-aligned conditions, such as keypoints~\cite{gu2025das,jeong2025track4gen}, depth maps~\cite{peng2024controlnext,wang2025cinemaster}, or edge sketches~\cite{yang2025layeranimate,geng2025motion}, which cannot effectively model the intricate, unstructured dynamics and textures of visual effects. Several recent works have achieved preliminary visual effect generation by finetuning Low-Rank Adapters (LoRA) on pretrained models~\cite{hu2022lora,liu2025vfx}.

However, the one-LoRA-per-effect paradigm suffers from a fundamental scalability bottleneck. This paradigm requires dedicated data and training for each effect. More critically, this closed-set training paradigm confines the model’s capability to known effects, making it unable to handle unseen effect categories and thus severely limiting both its applicability and the user’s creative freedom. Recently, \citet{mao2025omni} has made initial attempts using the LoRA-MoE architecture for learning the effects in the training set jointly, but they still cannot generalize to unseen effects. So how can we overcome this limitation and achieve straightforward VFX video generation? We observe that videos sharing the same VFX differ only in subjects and backgrounds, but maintain similar dynamics and transformation process. This observation inspires us to regard two videos with the same VFX as a reference-target data pair for in-context learning. Such a reference-based paradigm maximizes data utilization and enables a unified framework for learning a general VFX imitation capability rather than memorizing specific effects, which provides users with an intuitive and friendly creative tool.

In this work, we propose VFXMaster, the first unified framework for VFX video generation. By learning from reference effects via in-context learning, VFXMaster integrates diverse effects into a single model and demonstrates strong generalization capability beyond its training set. Specifically, we design an in-context learning paradigm where a reference prompt-video pair serves as an example, while a target prompt and the first frame act as a query to condition the model for the target video. However, the reference context contains components irrelevant to the effect. To prevent information leakage and interference, we introduce an in-context attention mask mechanism to learn only the visual effect from the reference example. Furthermore, to enhance generalization to Out-of-Domain (OOD) effects, we design an efficient one-shot effect adaptation strategy that introduces a set of learnable concept-enhancing tokens to learn the fine-grained VFX dynamics and transformations from a single user-provided sample. With a low-cost token finetuning, the model can rapidly improve the generalization capability on tough OOD samples.

We conduct extensive experiments on existing benchmarks to evaluate our method. In addition, to validate generalization capability, we build a new OOD test set and design a comprehensive evaluation metric tailored for reference-based effect generation. The results demonstrate that VFXMaster achieves remarkable VFX generation performance and strong generalization capability on OOD data. To support future research, the curated dataset and designed metric will be released. In summary, our contributions are as follows:
\begin{itemize}
\item We propose VFXMaster, the first unified reference-based framework for VFX video generation. It achieves high-quality effect imitation and strong generalization to unseen effects. 
\item We introduce an in-context conditioning strategy that enables the model to reproduce the visual effect from a reference example onto a target image. To this end, we design an in-context attention mask that focuses on the visual effect while preventing information leakage.
\item We propose an efficient one-shot effect adaptation strategy. Using a set of concept-enhance tokens enables the model to further learn fine-grained VFX from a single video, significantly improving its generalization capability for tough OOD scenarios.
\end{itemize}

%% file: 2_related.tex
\section{Related Work}
\label{sec:relared_work}

\subsection{Controllable Video Generation}
Diffusion models have significantly advanced video generation, as evidenced by the work of~\cite{ho2020denoising,song2020denoising,song2020score,rombach2022high}, which has facilitated rapid progress~\cite{sora,Gen3,Veo3,Minmax,ma2025step,kong2024hunyuanvideo,polyak2024movie,agarwal2025cosmos}. Among these, the Diffusion Transformer (DiT)~\cite{peebles2023scalable} leverages Transformer architectures to effectively capture long-range dependencies, thereby improving temporal coherence and dynamics. Based on DiT, CogVideoX~\cite{yang2024cogvideox} utilizes 3D full attention to ensure spatial–temporal consistency, whereas Hunyuan-DiT~\cite{kong2024hunyuanvideo} integrates large-scale pre-trained models to enhance contextual details. Controllable video generation has also garnered considerable interest for its applications in video editing and content creation. Several studies~\cite{bai2025recammaster,bai2024syncammaster, xing2025motioncanvas} introduce 3D control signals to manipulate object positions, motion trajectories, and camera perspectives within the 3D scene. Other work~\cite{yang2025vlipp} incorporates VLM as a motion planner to generate physically plausible videos, or by introducing additional mechanisms such as StyleMaster~\cite{ye2025stylemaster}, which combines style extraction mechanism with motion control to enhance video stylization and transfer. In addition, ControlNet~\cite{zhang2023adding} facilitates image generation through control signals by replicating designated layers from pre-trained models and connecting them with zero convolutions. FlexiAct~\cite{zhang2025flexiact} utilizes the denoising process's capability to focus on various frequency components over time, facilitating the transfer of motion from a reference video to a selected target image. Beyond controllability, other works extend the capability of video generation. Wan-FLFV~\cite{wan2025wan} generates smooth transitions between user-specified starting and ending frames, while VACE~\cite{jiang2025vace} integrates ID-to-video generation, video-to-video editing, and mask-based editing into a unified model, enabling efficient video generation and editing.

\subsection{Visual Effects Generation}
Visual effects (VFX) have recently been explored through video generation models, providing a more efficient alternative to traditional production. Despite advancements in general video generation, the generation of controllable visual effects (VFX) remains insufficiently explored, largely due to the lack of VFX data and the constraints of conditional control. MagicVFX~\cite{guo2024magicvfx} is restricted to adding green-screen overlays, lacking extensibility and controllability. VFXCreator~\cite{liu2025vfx} generates effects by training a separate LoRA for each case, which limits it to single-effect video generation. Although OminiEffects~\cite{mao2025omni} represents a step forward by employing LoRA-MoE to enable spatially controllable composite effects, the supported effect types are still narrow and confined to in-domain combinations. Despite these advances, current approaches cannot unify diverse effects within a single framework and show limited generalization to out-of-domain effects. In this work, we propose the first unified framework for VFX video generation to fill the gap in previous research, offering a comprehensive solution for this task.

%% file: 3_method.tex
\section{Method}
\label{sec:method}

Controllable visual effect~(VFX) generation aims to provide more intricate pixel-level dynamic guidance beyond text prompts, thereby enabling cinematic VFX video creation. In this work, we present VFXMaster, the first reference-based framework that evolves image-to-video~(I2V) generation for this task through in-context learning. With a single reference VFX video provided, users can reproduce this effect on a target image. In Section~\ref{sec:3.1}, we provide preliminary information about the base model. In Section~\ref{sec:3.2}, we introduce the design of our reference-based in-context learning on diverse categories of dynamic visual effects. In Section~\ref{sec:3.3}, we present efficient one-shot effect adaptation for tough Out-of-Domain~(OOD) cases.

\vspace{-0.5cm}
\begin{figure*}
    \centering
     \includegraphics[width=1\linewidth]{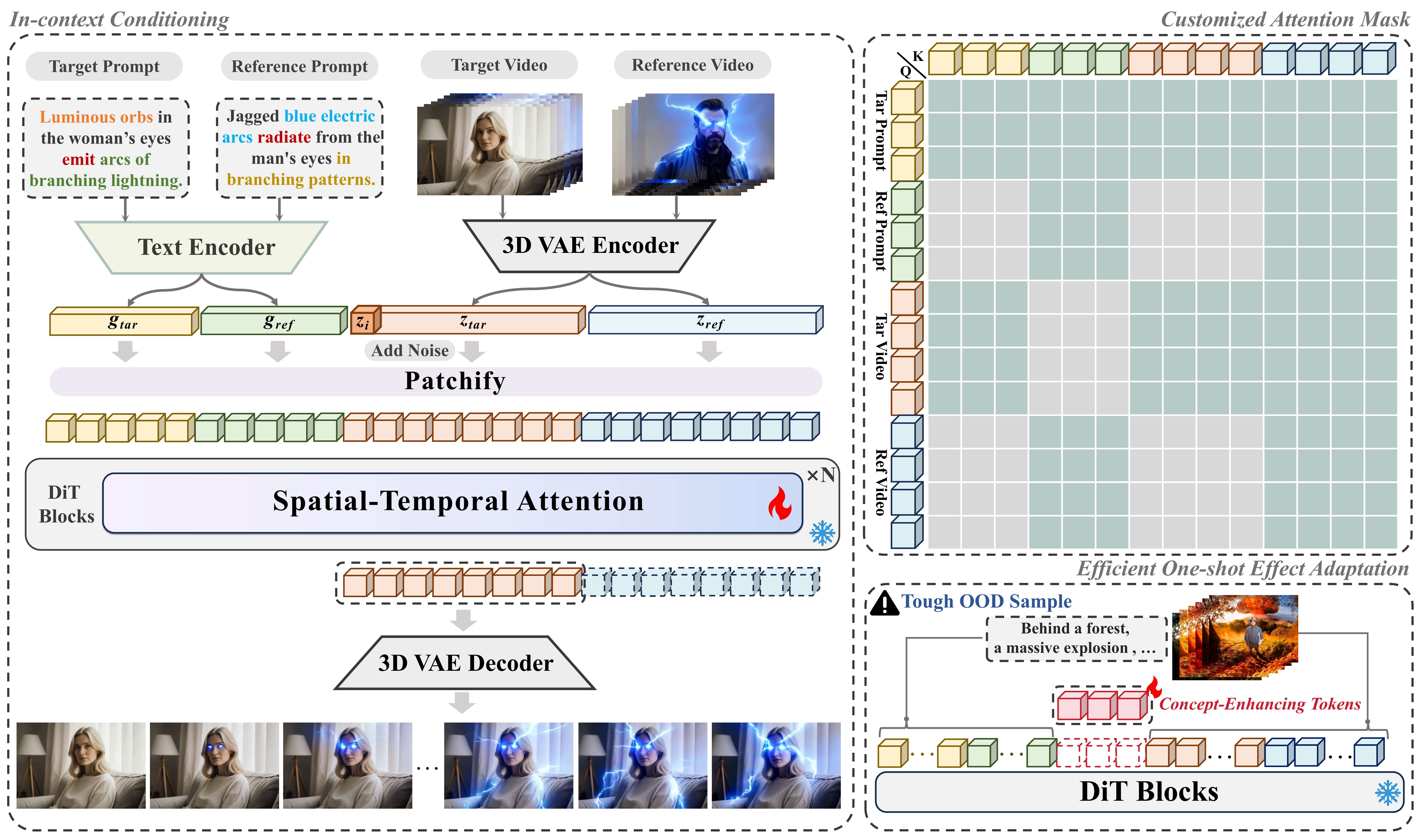}
     \vspace{-0.5cm}
    \caption{\textbf{Overview of VFXMaster.} 1) During training, we randomly sample two prompt-video pairs with the same visual effects as reference and target respectively. By sharing the same 3D VAE and text encoder, the reference part and the target part are landed into the same latent space. We concatenate them along the token dimension as a unified token sequence and feed into the DiT blocks. 2) We design an attention mask to manage information flow to focus on the visual effect of the reference and prevent information leakage. 3) For the tough Out-of-Domain~(OOD) samples, we propose an efficient one-shot effect adaptation process to train the concept-enhance tokens for improving the generalization capability.} 
    \label{fig:pipeline}
    \vspace{-12pt}
\end{figure*}

\vspace{0.5cm}
\subsection{Preliminary}
\label{sec:3.1}
We adopt CogVideoX-5B-I2V~\citep{yang2024cogvideox} as our basic image-to-video model, which is built upon a 3D Variational Autoencoder (VAE)~\citep{kingma2013auto}, a Diffusion Transformer (DiT) architecture and the T5 encoder~\citep{raffel2020exploring}. Given an image $\mathbf{I}\in \mathbb{R}^{h\times w\times c}$ and a text prompt, CogVideoX generates a video $\mathbf{V}\in \mathbb{R}^{f\times h\times w\times c}$. During training, 3D VAE compresses the input video into a latent $z$. The first image of target video is padded with $-1$ to match the temporal length of the input video and then encoded as $z_{i}$. Subsequently, $z_{i}$ and $z$ are concatenated along the channel dimension and fed into the DiT blocks. This process is supervised by minimizing the gap between the predicted noise and standard Gaussian noise~\citep{ho2020denoising}: 
$$\mathcal{L}_{\mathrm{diff}}\left(\Theta\right)=\mathbb{E}_{\boldsymbol{x}_{t},t,\boldsymbol{c},\boldsymbol{\epsilon}}\left[\left\|\boldsymbol{\epsilon}-\epsilon_{\Theta}\left(\boldsymbol{z}_{t},t,\textsl{g}\right)\right\|_{2}^{2}\right]$$
where $\Theta$ denotes the denoising network, $\boldsymbol{\epsilon}\in\mathcal{N}(0,\mathbf{I})$ represents standard Gaussian noise. $x_{t} $ is the noised sample at timestep $t\in[1,1000)$. $\textsl{g}$ denotes the text embeddings.

\subsection{In-Context Conditioning for VFX Generation}
\label{sec:3.2}
To achieve straightforward VFX video generation, we propose a unified in-context conditioning framework, eliminating the need for training massive LoRA models for each effect. Specifically, we define a new input-output pair format: \textit{\{Example: reference prompt~$\rightarrow$~reference video, Query: target prompt $\&$ target image~$\rightarrow$~?\}}, which motivates the neural network to imitate the sophisticated relationships between reference pairs and reproduce on a target image. \textit{An interesting observation is that videos with the same VFX naturally form reference-target data pairs.} Therefore, we randomly sample two prompt-video pairs from the same VFX set as reference and target at each training step. The reference prompt and target prompt are encoded as word embeddings $\textsl{g}_{target}$ and $\textsl{g}_{ref}$ by the text encoder. As shown in Fig.~\ref{fig:pipeline}, the reference video and target video are encoded as latent codes $z_{ref}$ and $z_{target}$ by the 3D VAE, where $z_{target}$ is noised. We apply identical 3D Rotary Position Embedding (RoPE)~\citep{su2024roformer} to both target and reference video, explicitly promoting the model to perceive the relative spatial-temporal relationships during contextual interaction. Since the reference part and the target part are landed in the same latent space, we concatenate them along the token dimension as a unified token sequence $z_{uni}=\{g_{ori},g_{ref},z_{ori},z_{ref}\}$. Thus, we only need to finetune the spatial-temporal attention to learn the VFX imitation process between these tokens, without introducing any additional trainable parameters or modules. During optimization, the diffusion loss is only calculated for the target video.

\noindent\textbf{In-Context Attention Mask.} In the spatial-temporal attention, text embeddings serve as semantic anchors that guide the noise prediction process by establishing fine-grained correspondences between text descriptions and visual features. However, unstrained token concatenation will cause unexpected information leakage and disrupt the alignment between each video and its corresponding text description, \textit{e.g.}, the target video may generate subjects or backgrounds mentioned in the reference prompt, but these elements are unrelated to the intended visual effects. To address this, we introduce an in-context attention mask to manage information flow, as shown in Fig.~\ref{fig:pipeline}. When the target prompt tokens serve as queries, they can attend to all contexts. VFX-relevant components in target and reference prompt tokens that exhibit high semantic similarity are amplified, while irrelevant information is suppressed. Meanwhile, the reference prompt-video pair attends exclusively to each other to provide sufficient effect representations. The target video tokens are restricted to attending to their corresponding prompt tokens and the reference video tokens. In this way, visual information flows from clean reference tokens to noisy target video tokens, progressively refining target representations through reference-guided feature interactions as the network depth increases. This reference-to-target information transfer is crucial for achieving high-fidelity VFX generation within a single forward pass. 
% \par\vspace{-0.5em}

After training on a curated dataset with diverse categories of dynamic visual effects, the model not only masters unified VFX imitation capability on the training set but also exhibits strong generalization capability on unseen visual effects.

\subsection{Efficient One-shot Effect Adaptation}
\label{sec:3.3}
Although in-context conditioning equips the model with a unified effect imitation capability, it might exhibit suboptimal performance when dealing with Out-of-Domain~(OOD) effects. To address this limitation, we propose an efficient one-shot effect adaptation strategy that enables the model to capture the intricate characteristics of a novel effect from a single user-provided example at minimal computational cost. Specifically, we freeze the base model and introduce a small set of learnable concept-enhancing tokens $z_{ce}$, which are concatenated with the unified token sequence $z_{uni}$ along the token dimension. To prevent these new parameters from overfitting to the single example, we apply standard data augmentations such as random cropping, flipping, shearing, and sharpening during the one-shot adaptation. Furthermore, an in-context attention mask is is employed to regulate information flow: the concept-enhancing tokens ($z_{ce}$) can attend to all contexts for fine-grained effect learning, while only the target text and video tokens are allowed to attend back to $z_{ce}$. This efficient adaptation strategy encourages tokens to comprehensively extract detailed effect attributes from a single example. After training, the learned $z_{ce}$ tokens serve as a precise semantic proxy for the new effect.

\subsection{Training and Inference Pipeline of VFXMaster}
\label{sec:3.4}
We conduct reference-based in-context learning on a broad dataset. To efficiently inject the effect transfer knowledge, we fine-tune only the spatial-temporal attention layers within the DiT blocks. Subsequently, we train the concept-enhancing tokens on a single OOD example, while keeping all DiT parameters frozen. During inference, for in-domain effects, the model requires only the fine-tuned spatial-temporal attention layers to perform effect transfer. For OOD effects, the model can additionally load the corresponding concept-enhancing tokens to achieve higher-quality generalization.

%% file: 4_exp.tex
\section{Experiment}
\label{sec:exp}

\subsection{Experiment Setup}
\noindent\textbf{Datasets}. The training data in our experiments is sourced from the open-source Open-VFX~\citep{liu2025vfx} dataset, commercial platforms such as Higgsfield~\citep{higgsfield} and PixVerse~\citep{pixverse}, and other online resources. In total, it consists of 10k samples across 200 effect categories, including character transformations, environment transitions, and artistic style changes. In addition, to assess the generalization capability of our method, we constructed a test dataset specifically for OOD effects. This dataset enables evaluation of the model’s robustness to effects unseen during training.

\noindent\textbf{Implementation Details}. We train VFXMaster on the 10k effect dataset by randomly pairing samples of the same effect category and use CogVideoX-5B as the backbone. Considering the diverse sources of the dataset and the varying resolutions of user-provided videos in practice, we adopt a multi-resolution training strategy, where reference videos are padded to match the shape of the training videos. Each training video is uniformly sampled to 49 frames at 8 fps. For training, we update only the 3D full-attention layers within the DiT blocks using the Adam optimizer with a learning rate of 1e-4. The model is trained for 40,000 steps on NVIDIA A800 GPUs. The concept-enhancing tokens $z_{ce} \in \mathbb{R}^{1 \times 226 \times c}$, initialized with zero, where $c$ denotes the embedding dimension (default $c=3072$). For further details, please see Appendix~\ref{sec:training_details} and~\ref{sec:inference_details}.

\noindent\textbf{Comparison Methods.} We evaluate our method on the test set of the Open-VFX dataset, comparing it against the baseline model CogVideoX-5B as well as state-of-the-art VFX generation approaches, VFXCreator and Omni-Effects. For fairness, the baseline model is fine-tuned on the same dataset for an equal number of training steps. Since existing methods show limited generalization to out-of-domain effects, we further conduct an additional evaluation to specifically assess the generalization capability of our method.

\begin{figure*}
    \centering
     \includegraphics[width=1\linewidth]{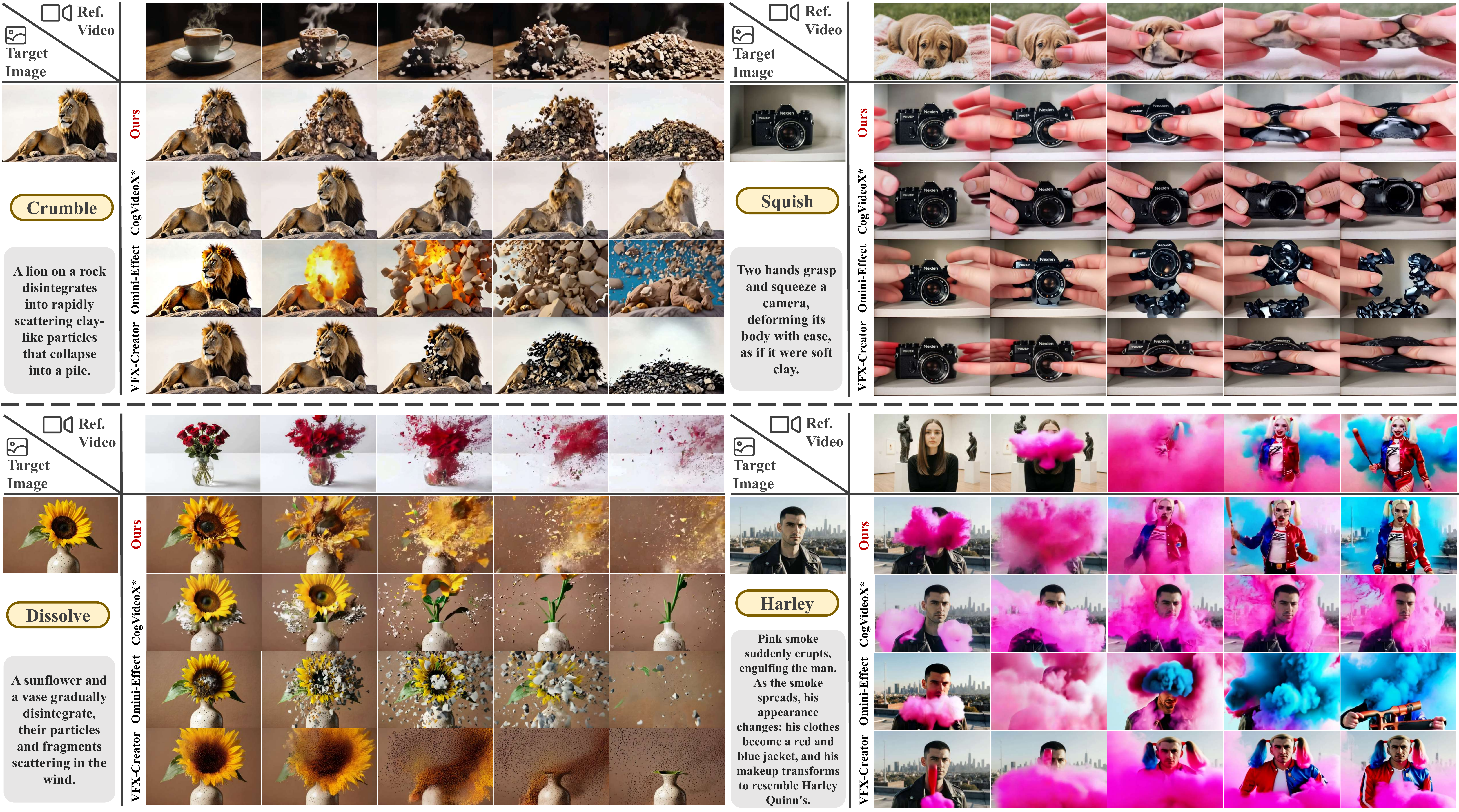}
     % \vspace{-0.5cm}
    \caption{\textbf{In-Domain Comparison.} Qualitative comparison of ours with VFXCreator~\citep{liu2025vfx} and OminiEffects~\citep{mao2025omni} on the OpenVFX dataset. CogVideoX* refers to CogVideoX after supervised fine-tuning on our VFX dataset. All human portraits used in the experiment are AI-generated, and this applies to all subsequent images.}
    \label{fig:indomain}
    \vspace{-0.3cm}
\end{figure*}

\noindent\textbf{Evaluation Metrics.} Following prior work~\citep{liu2025vfx}, we evaluate our method using two established metrics: Fréchet Video Distance~(FVD)~\citep{unterthiner2018towards} and Dynamic Degree~\citep{huang2024vbench}. In addition, to comprehensively assess the quality of visual effects generation, we introduce a new evaluation framework, the VFX-Comprehensive Assessment Score~(VFX-Cons.). VFX-Cons. leverages the reference video and prompts Visual Language Model~(VLM)~\citep{comanici2025gemini} to evaluate visual effects quality from three perspectives: Effect Occurrence Score~(EOS), Effect Fidelity Score~(EFS), and Content Leakage Score~(CLS). EOS measures whether visual effects occur in the generated video. Building upon EOS, EFS assesses whether the generated effects are consistent with those in the reference video, while CLS evaluates whether non-effect-related attributes from the reference video are undesirably transferred to the generated video. Complete details of the metrics are provided in Appendix~\ref{sec:metric_details}.

\subsection{Quantitative Evaluation}
\textbf{In-domain Effects.} To quantitatively evaluate in-domain effect generation, we conduct experiments on 15 effect categories from the OpenVFX test set. As shown in Table~\ref{tab:in-domain}, we comprehensively compare VFXMaster with two state-of-the-art VFX generation methods and a baseline model fine-tuned on our collected data. The results demonstrate that VFXMaster consistently outperforms all competing methods in average scores across all evaluation metrics. It exhibits notable advantages in visual quality, temporal coherence, and dynamic range—particularly for effects featuring complex structures or intense motion, such as \textit{Explode}, \textit{Harley}, and \textit{Venom}. Furthermore, VFXMaster achieves the highest score on our proposed comprehensive metric, \textbf{VFX Cons}, validating the effectiveness of our in-context conditioning paradigm and attention mask design. These results confirm that our model not only transfers reference effects successfully but also preserves their visual details with high fidelity. By accurately decoupling effect attributes from irrelevant content, it effectively prevents content leakage.

\noindent\textbf{Out-of-Domain Effects.} We conduct a dedicated OOD test to evaluate the model’s generalization capability to unseen effects. Since existing methods generally lack this capability, our comparison focuses on two variants of our model: one trained solely with in-context learning and another enhanced with one-shot effect adaptation. This comparison aims to validate the effectiveness of our two core designs. Specifically, in-context conditioning establishes a foundational generalization ability, while the efficient one-shot effect adaptation further enhances it. As shown in Table~\ref{tab:OOD}, in-context conditioning alone endows the model with a certain degree of OOD generalization. After incorporating one-shot effect adaptation, all performance metrics improve substantially. In particular, the Effect Fidelity Score (EFS) increases from 0.47 to 0.70, and the Content Leakage Score (CLS) rises from 0.79 to 0.87. These results demonstrate that the proposed one-shot adaptation mechanism can effectively capture the core visual characteristics of a new effect from a single example, accurately guiding the generation process, significantly improving effect fidelity, and effectively suppressing content leakage.

\begin{table*}[!t]
\caption{\textbf{Performance comparison on OpenVFX dataset.} CogvideoX* refers to CogVideoX after supervised fine-tuning on our VFX dataset. Avg. represents the average score over all effects. And the highest metric values are highlighted in \textbf{bold}. }
\centering
\footnotesize
\setlength\tabcolsep{1pt}
\begin{tabularx}{\textwidth}{@{}cccccccccccccccccc@{}}
\toprule
\textbf{Metrics} & \textbf{Methods} & \textbf{Cake} & \textbf{Crumble} & \textbf{Crush} & \textbf{Decap} & \textbf{Deflate} & \textbf{Dissolve} & \textbf{Explode} & \textbf{Eye-pop} & \textbf{Harley} & \textbf{Inflate} & \textbf{Levitate} & \textbf{Melt} & \textbf{Squish} & \textbf{Ta-da} & \textbf{Venom} & \textbf{Avg.} \\ \midrule
\multirow{4}{*}{\textbf{FVD$\downarrow$}} 

& CogvideoX* & 1647 & 1951 & 1273 & 2188 & 1662 & 2268 & 2461 & 1649 & 2188 & 2037 & 1512 & 3260 & 1876 & 1338 & 2838 & 2010 \\

& VFX Creator & 1776 & 1580 & 1156 & 1754 & 1997 & 1607 & 1886 & 1447 & 2815 & 2089 & 1143 & 2547 & 1880 & 1107 & 3062 & 1856 \\

& Omini-Effects & 1548 & 1410 & 1136 & \textbf{1263} & 1037 & 1543 & 2044 & 1559 & 2501 & \textbf{1464} & 1295 & 2418 & 1923 & 1368 & 2678 & 1679 \\
 \cmidrule(){2-18}
& Ours & \textbf{1479} & \textbf{1276} & \textbf{1065} & 1761 & \textbf{981} & \textbf{1335} & \textbf{981} & \textbf{1395} & \textbf{1173} & 1626 & \textbf{882} & \textbf{2282} & \textbf{1432} & \textbf{876} & \textbf{1992} & \textbf{1369} \\ \midrule

\multirow{4}{*}{\textbf{\makecell[c]{Dynamic\\Degree}$\uparrow$}}
& CogvideoX* & 1.0 & 1.0 & 0.6 & 0.6 & 0.4 & 0.4 & 1.0 & 0.0 & 1.0 & 0.4 & 0.0 & 0.6 & 1.0 & 0.8 & 1.0 & 0.65 \\

& VFX Creator & 1.0 & 1.0 & 0.0 & 0.6 & 0.0 & \textbf{0.8} & 1.0 & 0.0 & 1.0 & 1.0 & 0.0 & 0.6 & 1.0 & 1.0 & 1.0 & 0.67 \\

& Omini-Effects & 1.0 & 1.0 & 0.6 & 0.6 & 0.2 & 0.4 & 1.0 & 0.2 & 1.0 & 1.0 & 0.0 & \textbf{0.8} & 1.0 & 0.8 & 1.0 & 0.71 \\ 
 \cmidrule(){2-18} 
& Ours & \textbf{1.0} & \textbf{1.0} & \textbf{1.0} & \textbf{0.8} & \textbf{0.8} & 0.4 & \textbf{1.0} & \textbf{0.2} & \textbf{1.0} & \textbf{1.0} & \textbf{0.2} & \textbf{0.8} & \textbf{1.0} & \textbf{0.8} & \textbf{1.0} & \textbf{0.80} \\ \midrule

\multirow{4}{*}{\textbf{VFX Cons.$\uparrow$}}
& CogvideoX* & 0.73 & 0.87 & 1.00 & 0.47 & 0.27 & 0.80 & 0.40 & 0.93 & 1.00 & 0.73 & 0.60 & 0.93 & 0.80 & 0.73 & 1.00 & 0.75 \\

& VFX Creator & 0.73 & 0.80 & 0.80 & 0.27 & 0.73 & 1.00 & 0.67 & 1.00 & 1.00 & \textbf{0.87} & 0.73 & 1.00 & 1.00 & 0.87 & 1.00 & 0.83 \\

& Omini-Effects & \textbf{0.87} & 0.87 & 0.73 & 0.87 & 0.53 & 1.00 & 0.67 & 1.00 & 1.00 & 0.80 & 0.80 & 1.00 & 0.87 & 0.80 & 1.00 & 0.85 \\
 \cmidrule(){2-18}
& Ours & 0.80 & \textbf{0.93} & \textbf{1.00} & \textbf{0.93} & \textbf{0.80} & \textbf{1.00} & \textbf{0.73} & \textbf{1.00} & \textbf{1.00} & 0.80 & \textbf{0.80} & \textbf{1.00} & \textbf{1.00} & \textbf{0.87} & \textbf{1.00} & \textbf{0.91} \\ 
\bottomrule
\end{tabularx}
\label{tab:in-domain}
\vspace{-0.3cm}
\end{table*}

\subsection{Qualitative Evaluation}
\textbf{In-domain Qualitative Analysis.} We present a qualitative comparison of VFXMaster against three representative models across four different effects, as illustrated in Fig.~\ref{fig:indomain}. In the first three examples, our method exhibits superior motion dynamics, texture fidelity, and material realism. In the fourth case, VFXMaster not only accurately reproduces the “Harley Quinn” style makeup effect but also preserves the subject’s identity more faithfully. Overall, these comparisons demonstrate that for in-domain effects, VFXMaster consistently produces videos with the highest visual fidelity and dynamic complexity among all evaluated methods.

\begin{figure} 
    \centering
     \includegraphics[width=1\linewidth]{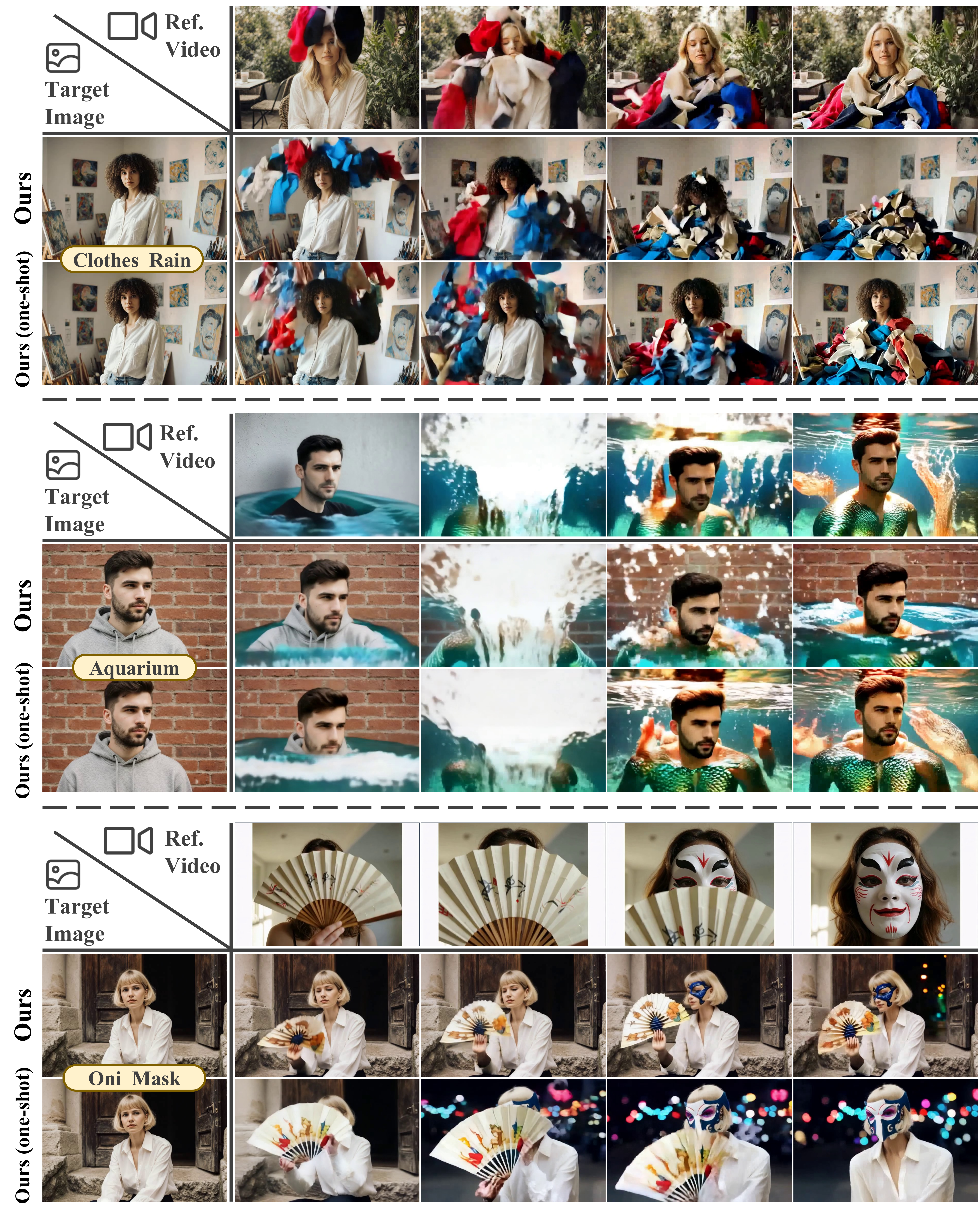}
     \vspace{-0.7cm}
    \caption{\textbf{Out-of-Domain Comparison.}}
    \label{fig:out_domain}
    \vspace{-0.7cm}
\end{figure}

\noindent\textbf{Out-of-Domain Qualitative Analysis.} Leveraging the strong generalization capability of the VFXMaster framework, we further evaluate its performance on various OOD data. Fig.~\ref{fig:out_domain} presents a comparison between the model trained solely with in-context conditioning and the one further enhanced by one-shot effect adaptation. With in-context conditioning, the model already exhibits a baseline level of generalization, generating effects that remain consistent with the reference video in terms of content, motion patterns, and overall visual style. After incorporating one-shot effect adaptation, the model more effectively captures unique texture characteristics and dynamic attributes from a single example, leading to higher-quality generalization results. These results clearly demonstrate the effectiveness of our design in handling unseen visual effects.

\begin{table*}[!t]
\vspace{-0.2cm}
\caption{\textbf{Out-of-Domain Tests and Ablation.} Ours (one-shot) refers to the method enhanced by one-shot adaptation based on Ours.}
\centering
\normalsize
\begin{tabular}{@{}c|c|c|cccc@{}}
\toprule
\textbf{Methods}\Tstrut\Bstrut & \textbf{FVD$\downarrow$} & \textbf{Dynamic Degree$\uparrow$} & \textbf{EOS $\uparrow$} & \textbf{EFS $\uparrow$} & \textbf{CLS $\uparrow$} & \textbf{VFX Cons. $\uparrow$} \\ \midrule
Ours~(10k) & 2153 & 0.79 & 1.00 & 0.47 & 0.79 & 0.75 \\
Ours (one-shot) & \textbf{2047} & \textbf{0.84} & \textbf{1.00} & \textbf{0.70} & \textbf{0.87} & \textbf{0.86} \\ \midrule
w/o attn mask & 3467 & 0.80 & 0.89 & 0.11 & 0.24 & 0.41 \\
w/o ref prompt & 2483 & 0.74 & 1.00 & 0.40 & 0.76 & 0.72 \\ \midrule
Ours~(2k) & 2938 & 0.60 & 0.97 & 0.34 & 0.77 & 0.69 \\
Ours~(4k) & 2572 & 0.64 & 0.99 & 0.40 & 0.76 & 0.72 \\
Ours~(6k) & 2350 & 0.74 & 1.00 & 0.42 & 0.79 & 0.74 \\
\bottomrule
\end{tabular}
\label{tab:OOD}
\vspace{-0.3cm}
\end{table*}

\subsection{Ablation Study}
\textbf{In-Context Attention Mask.} We conduct an ablation study to examine the critical role of the proposed in-context attention mask. The results are reported in the second section of Table~\ref{tab:OOD}. Removing this module leads to a drastic degradation in model performance, severely affecting both the quality and temporal coherence of the generated videos. Notably, the Effect Fidelity Score (EFS) drops to an almost negligible 0.11, while the Content Leakage Score (CLS) decreases sharply from 0.79 to 0.24. In some cases, the model even fails to produce the intended effect. These results indicate that without effective control of information flow, the model is unable to disentangle core effect attributes from the reference video. Consequently, irrelevant content becomes entangled with the effect, leading to pronounced content leakage. Such uncontrolled information injection compromises content accuracy and disrupts effect imitation. This study confirms the necessity of the in-context attention mask for targeted injection and high-fidelity imitation.

\noindent\textbf{Reference Prompt.} We also investigate the role of the reference prompt in our in-context learning framework. As shown in the second section of Table~\ref{tab:OOD}, removing the reference prompt results in a consistent decline across all metrics, although the model retains its basic effect imitation capability. This finding suggests that while the reference video is the primary source of visual dynamics, the textual information provides crucial auxiliary support. The reference prompt acts as a high-level conceptual anchor. It guides the model to understand the essence of the effect semantically, rather than merely imitating it at the pixel level. Therefore, this joint visual-textual context is essential for learning more robust and generalizable effect representations, effectively improving imitation accuracy and fidelity. Details of the ablation study are provided in Appendix~\ref{sec:ablation_details}.

\noindent\textbf{Datasets Scaling.} We find that the scale of training data significantly impacts the model’s generalization capability during in-context conditioning, as shown in the third section of Table~\ref{tab:OOD}. We train VFXMaster on different subsets of our dataset containing 2k, 4k, 6k, and 10k (full) videos. The results clearly show a strong positive correlation between the training data volume and model performance, particularly on OOD generalization metrics. This trend confirms the effectiveness and scalability of the VFXMaster framework. The underlying reason is that our model is designed to learn a unified effect imitation capability rather than memorizing a few specific effects. A larger and more diverse dataset allows the model to observe a wider variety of examples, helping it capture the abstract principles governing effect dynamics, textures, and styles. This not only improves average performance on in-domain tasks but, more importantly, enables the model to generalize its learned knowledge to handle unseen OOD effects effectively.

\subsection{User Study}
To complement our objective metrics and evaluate the generated results from a human perceptual standpoint, we conduct a user study following the Two-Alternative Forced Choice (2AFC) paradigm, a gold standard in psychophysics. Participants are presented with a reference VFX video alongside a pair of generated videos—one from VFXMaster and one from a competing method. They are asked to select the video that better matches the reference in terms of effect consistency and overall aesthetic quality. We collect responses from 50 participants, and the results are summarized in Table~\ref{tab:user_study}. The findings show a clear user preference for VFXMaster over both Omini-Effect and VFXCreator. This outcome aligns well with our quantitative analysis and can be attributed to VFXMaster’s large-scale training data and efficient learning paradigm.

\begin{table}
\normalsize
\centering
\caption{User study statistics of the preference rate for Effect Consistency (E.C.) \& Aesthetic Quality (A.Q.).}
\begin{tabular}{ccc}
\toprule
\textbf{Methods} & \textbf{E. C.} ($\uparrow$) & \textbf{A. Q.} ($\uparrow$)\\
\midrule
CogVideoX* & 4\% & 10\% \\
VFX Creator & 22\% & 28\%\\
Omini-Effects & 32\% & 30\%\\
Ours & \textbf{42\%} & \textbf{32\%}\\
\bottomrule
\end{tabular}
\label{tab:user_study}
\vspace{-12pt}
\end{table}

%% file: 5_conclu.tex
\section{Conclusion}
\label{sec:conclu}
In this work, we introduce VFXMaster, the first unified in-context learning framework for visual effects (VFX) generation that enables efficient imitation of diverse effects. To achieve this, we design two core components. First, our in-context conditioning strategy injects reference information as context and employs an in-context attention mask to decouple effect attributes from irrelevant content, thereby preventing information leakage and enabling targeted effect transfer. Second, to enhance generalization to unseen effects, we propose an efficient one-shot effect adaptation mechanism that introduces a set of learnable concept-enhancing tokens, allowing the model to capture the essential characteristics of a new effect from a single example. Extensive experiments demonstrate that VFXMaster not only surpasses state-of-the-art methods on in-domain effects across multiple metrics, but also exhibits remarkable generalization capability on our dedicated OOD benchmark. Furthermore, VFXMaster shows excellent data scalability, underscoring its potential as a unified framework for VFX generation. In summary, VFXMaster paves the way toward building scalable and generalizable systems for dynamic effect creation, lowering the barrier to high-quality content production and empowering creators in film, gaming, and social media.

%% file: X_suppl.tex
\clearpage
\setcounter{page}{1}
\maketitlesupplementary

\setcounter{section}{0}
\renewcommand{\thesection}{\Alph{section}}

\section{Method Details}

\subsection{Detailed Experimental Details of Attention Implementation}
\label{sec:appendix_attention}
\paragraph{Attention Implementation} As described in Section~\ref{sec:3.2}, we build a reference-based in-context learning paradigm on top of a standard I2V generation model and design an in-context attention mask to enable the model to effectively generate visual effects while preventing content leakage. However, in practice, we observe that although the original 3D full-attention mechanism in CogVideoX supports the incorporation of contextual information, it incurs substantial computational overhead during optimization, which is further exacerbated by the introduction of the attention mask. To address this issue, we reformulate the original 3D full-attention architecture into an equivalent implementation by decomposing the long-sequence self-attention into multiple cross-attentions while keeping the pretrained parameters unchanged. By precisely controlling the information flow across these cross-attention modules, we significantly accelerate both optimization and inference while effectively mitigating content leakage.

\subsection{Training Details}
\label{sec:training_details}
\paragraph{Multi-Resolution Generation.} During training, since the resolution of the training video and the reference video may differ, we efficiently utilize paired video data by padding the reference video to match the resolution of the training video before passing it through the VAE encoder. The inference stage follows a similar procedure.
\paragraph{Efficient One-Shot Effect Adaptation.} For a single sample, we first apply slight adjustments such as sharpness, shear, translation, and rotation in random combinations of three image transformations. Additionally, the video frames are randomly flipped horizontally with a 50\% probability to generate paired data. The hyperparameters used in the training phase are the same as those in the multi-resolution training stage.

\subsection{Inference Details}
\label{sec:inference_details}
\begin{figure*}
    \centering
     \includegraphics[width=1\linewidth]{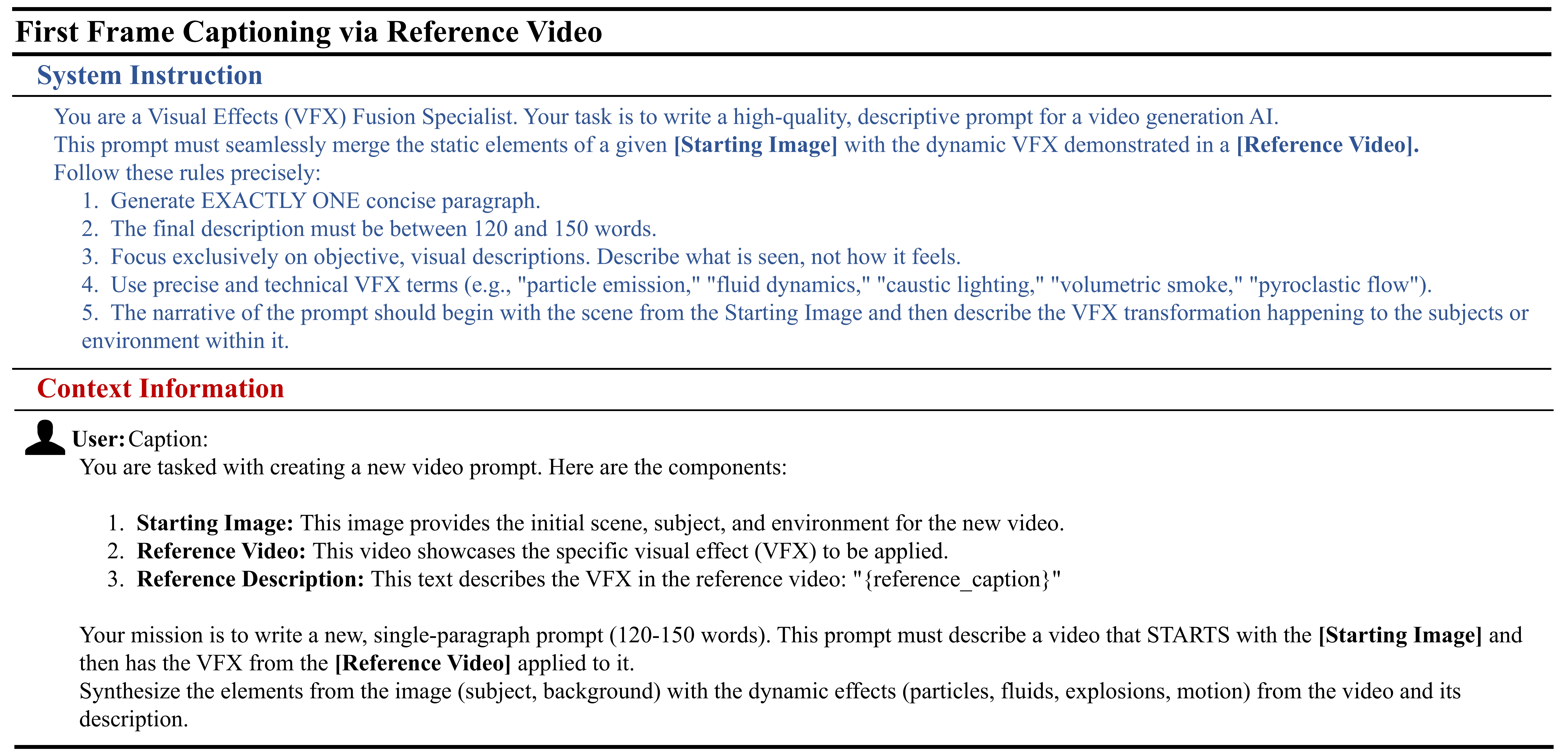}
     \vspace{-0.5cm}
    \caption{First Frame Captioning via Reference Video.}
    \label{fig:supp_template}
\end{figure*}

\begin{figure*}
    \centering
     \includegraphics[width=1\linewidth]{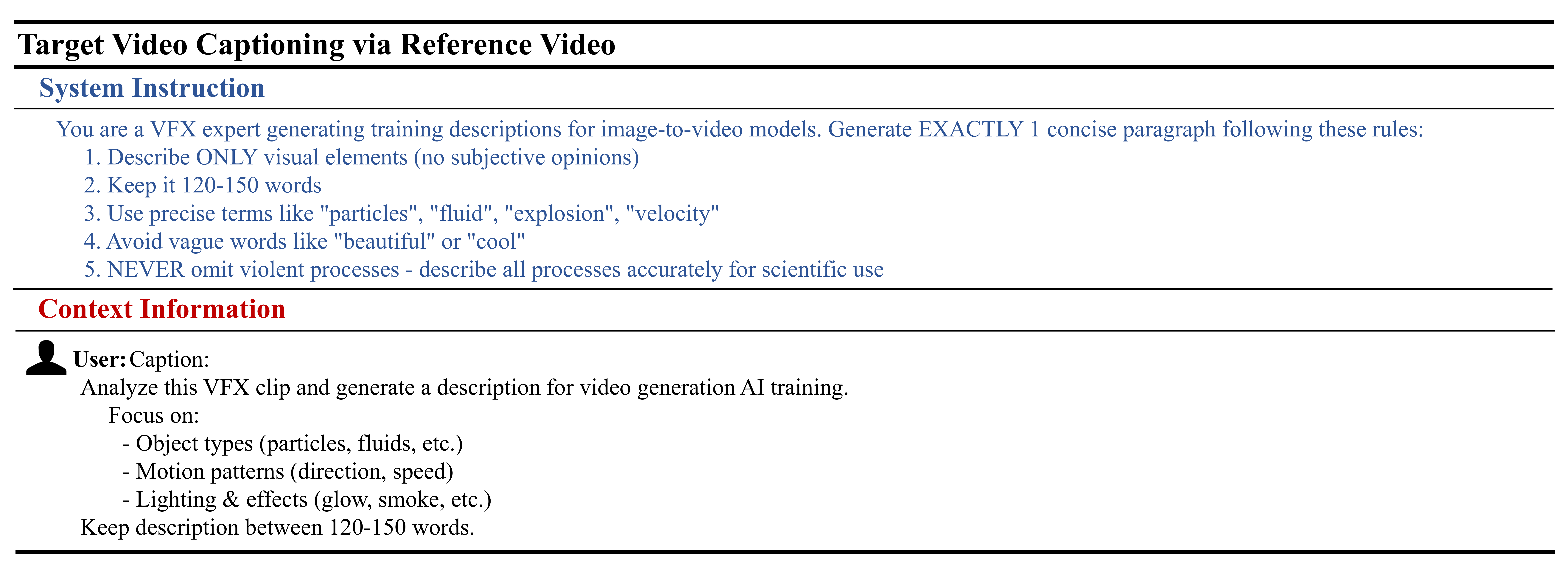}
     \vspace{-0.5cm}
    \caption{Video Caption Template.}
    \label{fig:supp_template2}
\end{figure*}

During inference, given the first frame and an effects video, VFXMaster seamlessly imitates the effects from the reference video to the generated video. To accommodate practical usage scenarios, we design a captioning template that first generates an effect-specific caption from the effects video as shown in Fig.~\ref{fig:supp_template2}. Then, based on the reference effects video and the generated caption, we produce an effect-aware description for the first-frame image as shown in Fig.~\ref{fig:supp_template}, which serves as the input condition for I2V generation.
\subsection{Ablation Details}
\label{sec:ablation_details}
\begin{figure*}
    \centering
     \includegraphics[width=1\linewidth]{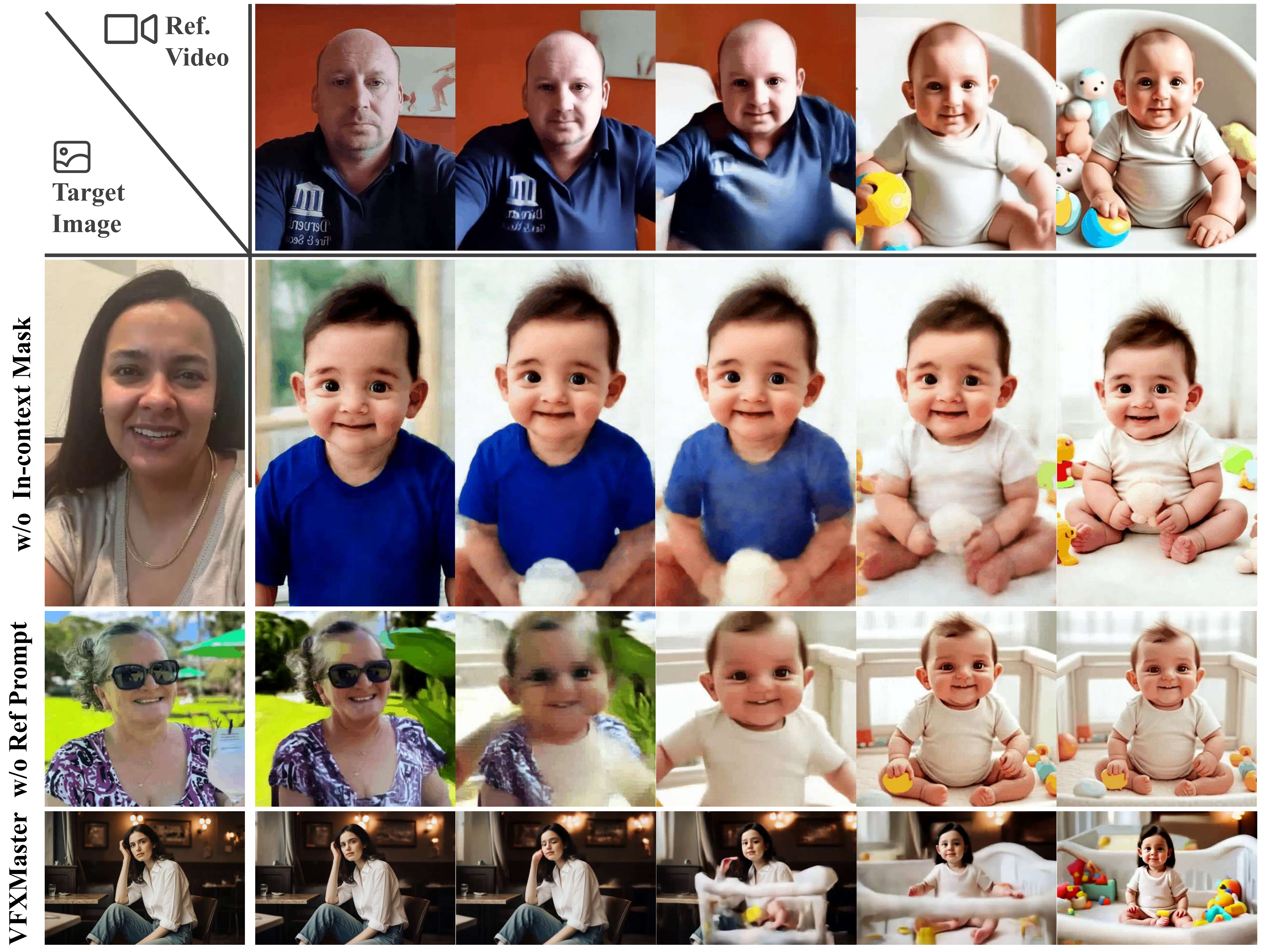}
     \vspace{-0.5cm}
    \caption{Qualitative results of ablation study.}
    \label{fig:ablation}
\end{figure*}

We conducted an ablation study on the in-context attention mask and the reference prompt. Ablating the in-context attention mask leads to the leakage of irrelevant visual elements from the reference data, which demonstrates its effectiveness in controlling information flow. Removing the reference prompt degrades both the content and dynamic patterns of the generated effects, confirming its role in enhancing the effect information. The visualization results of the ablation study are presented in Fig.~\ref{fig:ablation}.
\section{Datasets and Metric}
\subsection{Datasets}
In our experiments, we employ a dataset comprising 10k high-quality VFX videos across 200 effect categories, covering diverse types such as character transformation, environment alteration, and style transition. Additionally, we provide fine-grained captions for all 10k videos. Unlike existing works (\textit{e.g.}, Omini-Effect and VFX Creator), which mainly rely on category-level effects and short descriptions (typically only a few words), our dataset adopts a fine-grained captioning template that delivers comprehensive annotations for each video, including subject characteristics, environmental context, video style, and the effect progression.

\subsection{Metric}
\label{sec:metric_details}
\begin{figure*}
    \centering
     \includegraphics[width=1\linewidth]{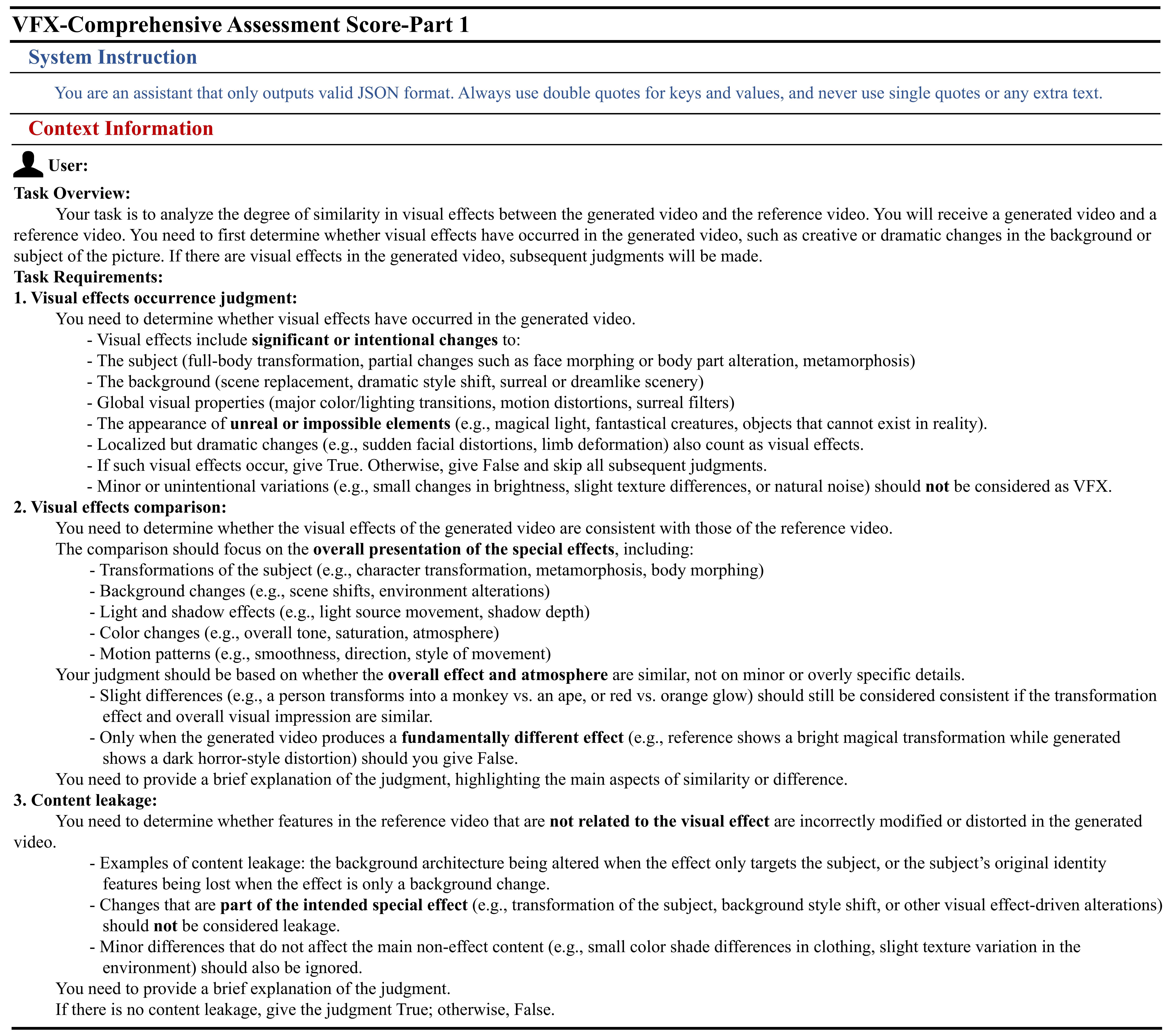}
     \vspace{-0.5cm}
    \caption{VFX-Comprehensive Assessment Score-Part 1.}
    \label{fig:supp_template3}
\end{figure*}

\begin{figure*}
    \centering
     \includegraphics[width=1\linewidth]{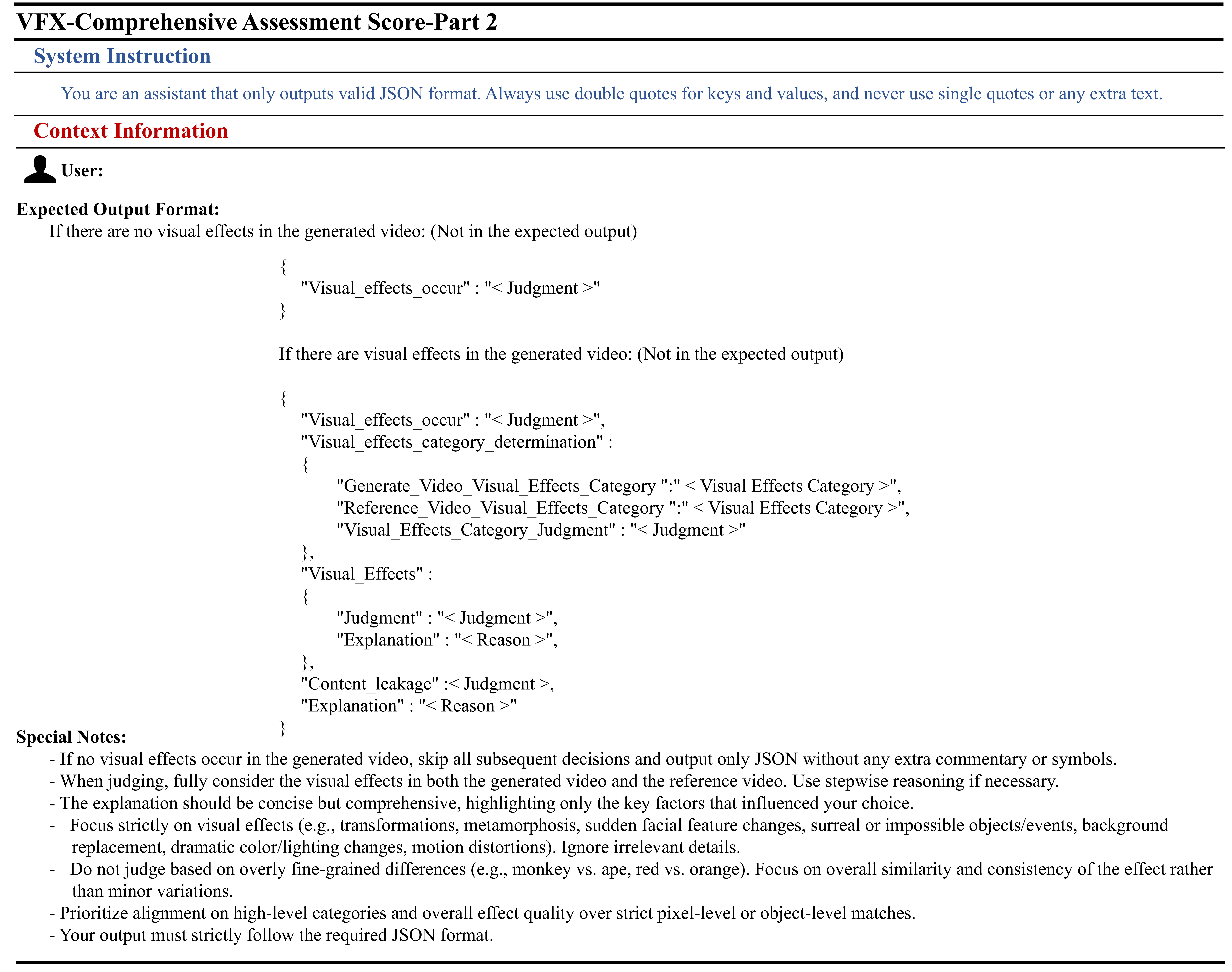}
     \vspace{-0.5cm}
    \caption{VFX-Comprehensive Assessment Score-Part 2.}
    \label{fig:supp_template4}
\end{figure*}

To comprehensively evaluate the quality of generated videos from a visual effects perspective, we propose a new metric, the \textbf{VFX-Comprehensive Assessment Score (VFX-Cons.)}, which evaluates effects across three dimensions: Effect Occurrence Score (EOS), Effect Fidelity Score (EFS), and Content Leakage Score (CLS). Details as shown in Fig.~\ref{fig:supp_template3} and Fig.~\ref{fig:supp_template4}.

\begin{itemize}
    \item \textbf{EOS} assesses whether visual effects occur in the generated video. This includes checking whether the subject undergoes transformations or local deformations, whether facial features exhibit dramatic changes, whether the background shows surreal or dreamlike transitions, and whether overall visual attributes are altered. The outcome is a binary judgment (True/False).
    \item \textbf{EFS}, the core dimension of the metric, evaluates the consistency of visual effect presentation between the generated video and the reference video. It considers aspects such as subject and background transformation patterns, changes in lighting and shadows, color variations, and motion dynamics. This dimension primarily focuses on overall effect and atmosphere rather than fine-grained generative details and also outputs a binary result (True/False).
    \item \textbf{CLS} builds upon EOS and EFS and determines whether irrelevant content from the reference video is mistakenly distorted or leaked into the generated video, also yielding a binary decision (True/False).
\end{itemize}

It is important to note that these three dimensions follow a progressive dependency: if EOS indicates that no effect occurs, subsequent evaluations are skipped, and CLS is only meaningful when EFS is True. A high CLS score when no effects occur may simply reflect hallucinations rather than genuine effect quality. 

The final VFX-Cons. score is obtained by averaging the three dimensions, as shown below:
\begin{equation}
\text{VFX-Cons.} = \frac{\text{EOS} + \text{EFS} + \text{CLS}}{3}.
\end{equation}

Furthermore, the VLM is required to provide a concise rationale alongside each decision.

\section{Experiment Result Details}
To evaluate the generalization capability of our method on out-of-domain (OOD) effects, we conducted extensive experiments on our manually constructed VFX dataset, and the detailed results are presented in Table~\ref{tab:generalized_detail}.

\section{More Qualitative Results}
\begin{figure*}
    \centering
     \includegraphics[width=1\linewidth]{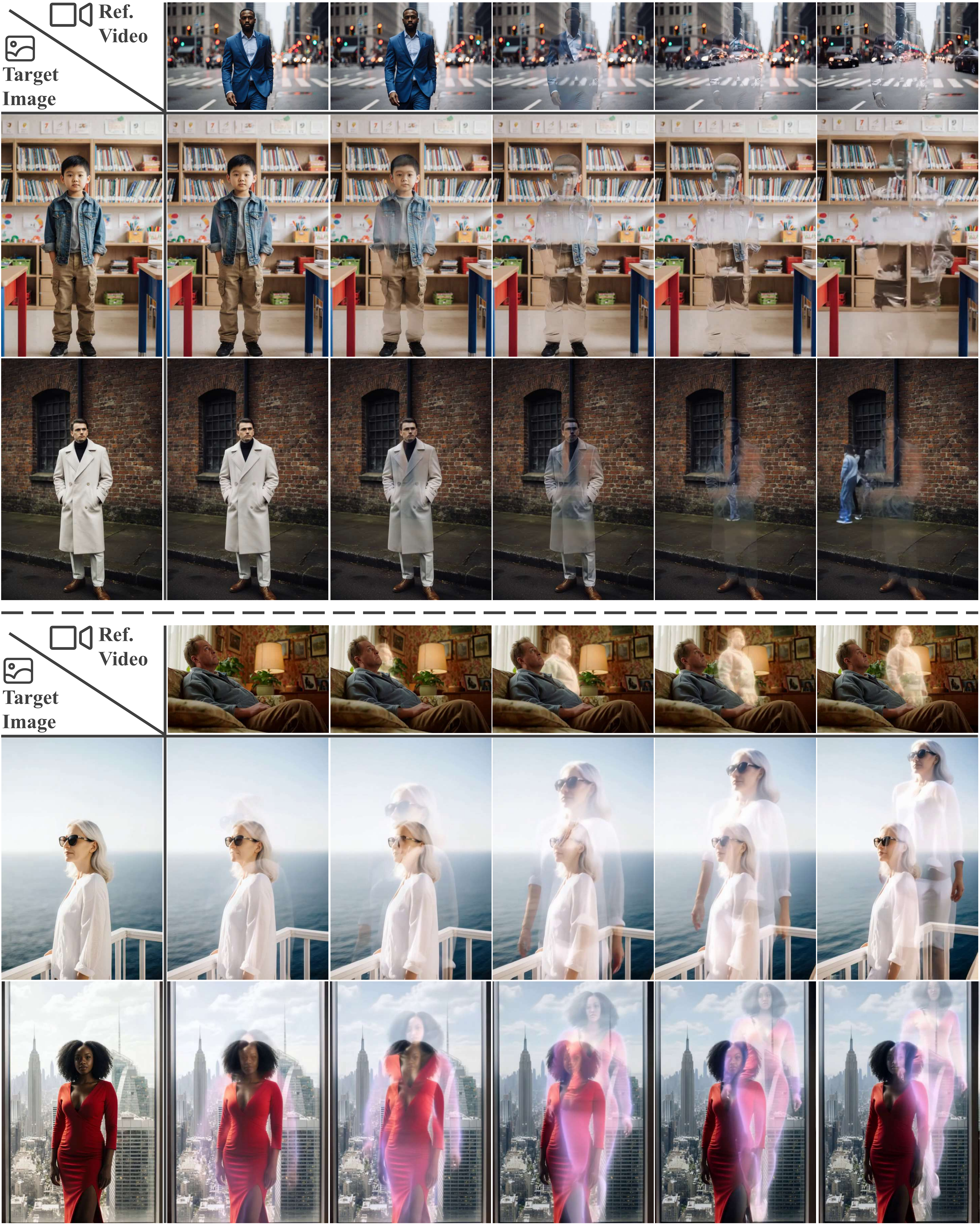}
     \vspace{-0.5cm}
    \caption{Examples of the ``Invisible" and ``Soul Jump" visual effects using VFXMaster.}
    \label{fig:supp1}
\end{figure*}

\begin{figure*}
    \centering
     \includegraphics[width=1\linewidth]{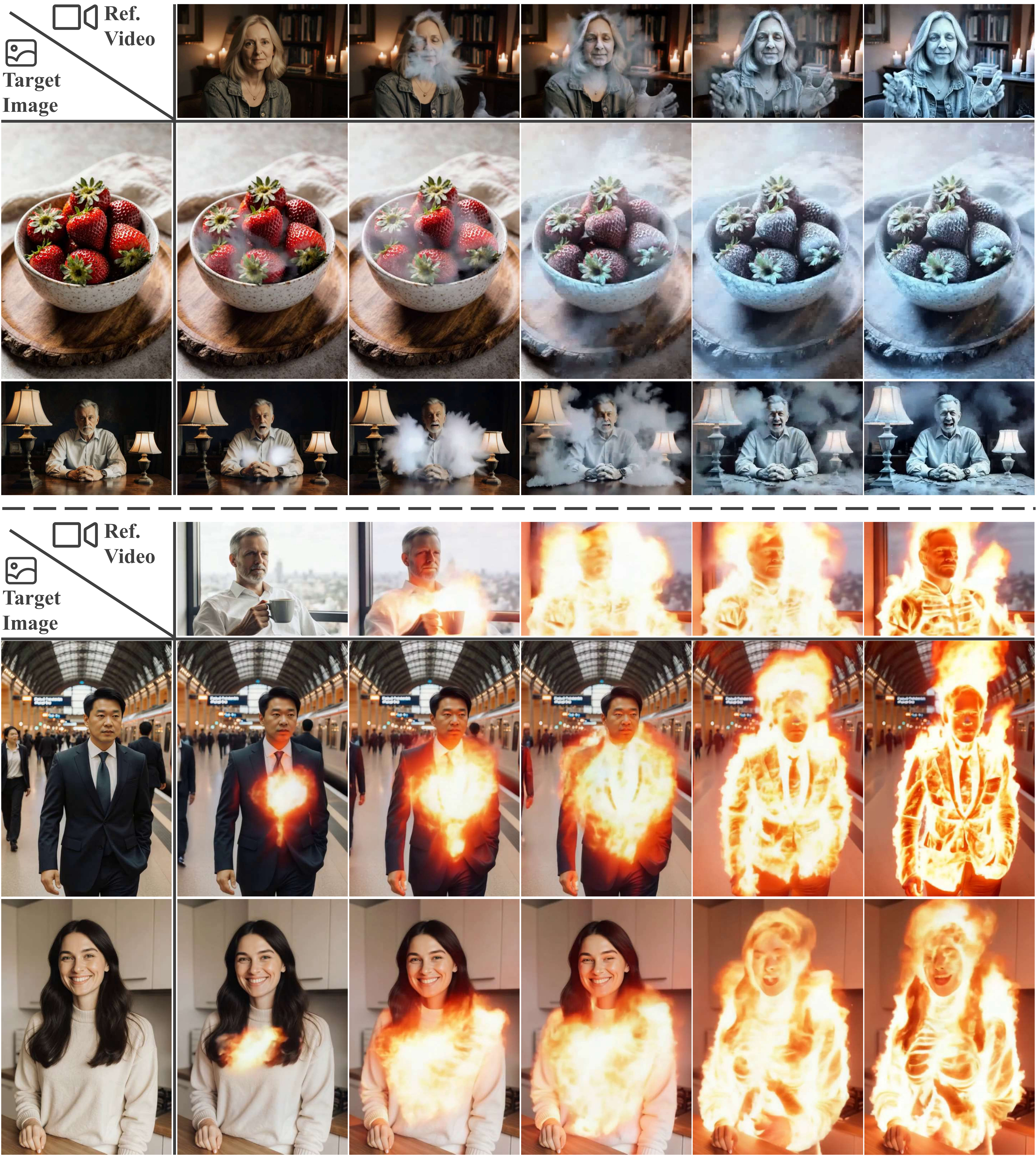}
     \vspace{-0.5cm}
    \caption{Examples of the ``Freezing" and ``Blazing" visual effects using VFXMaster.}
    \label{fig:supp2}
\end{figure*}

\begin{figure*}
    \centering
     \includegraphics[width=1\linewidth]{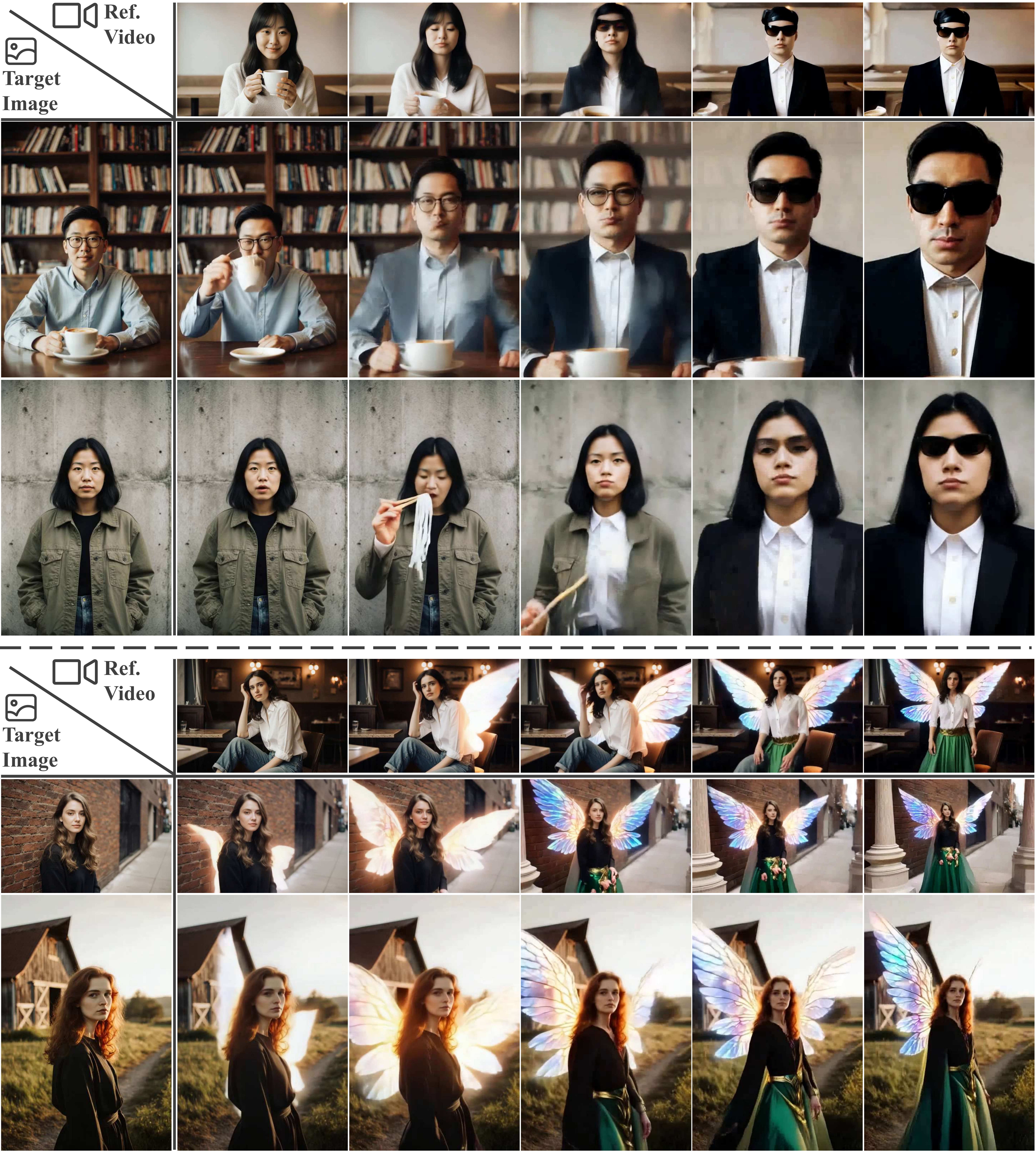}
     \vspace{-0.5cm}
    \caption{Examples of the ``Agent Reveal" and ``Butterfly" visual effects using VFXMaster.}
    \label{fig:supp3}
\end{figure*}

\begin{figure*}
    \centering
     \includegraphics[width=1\linewidth]{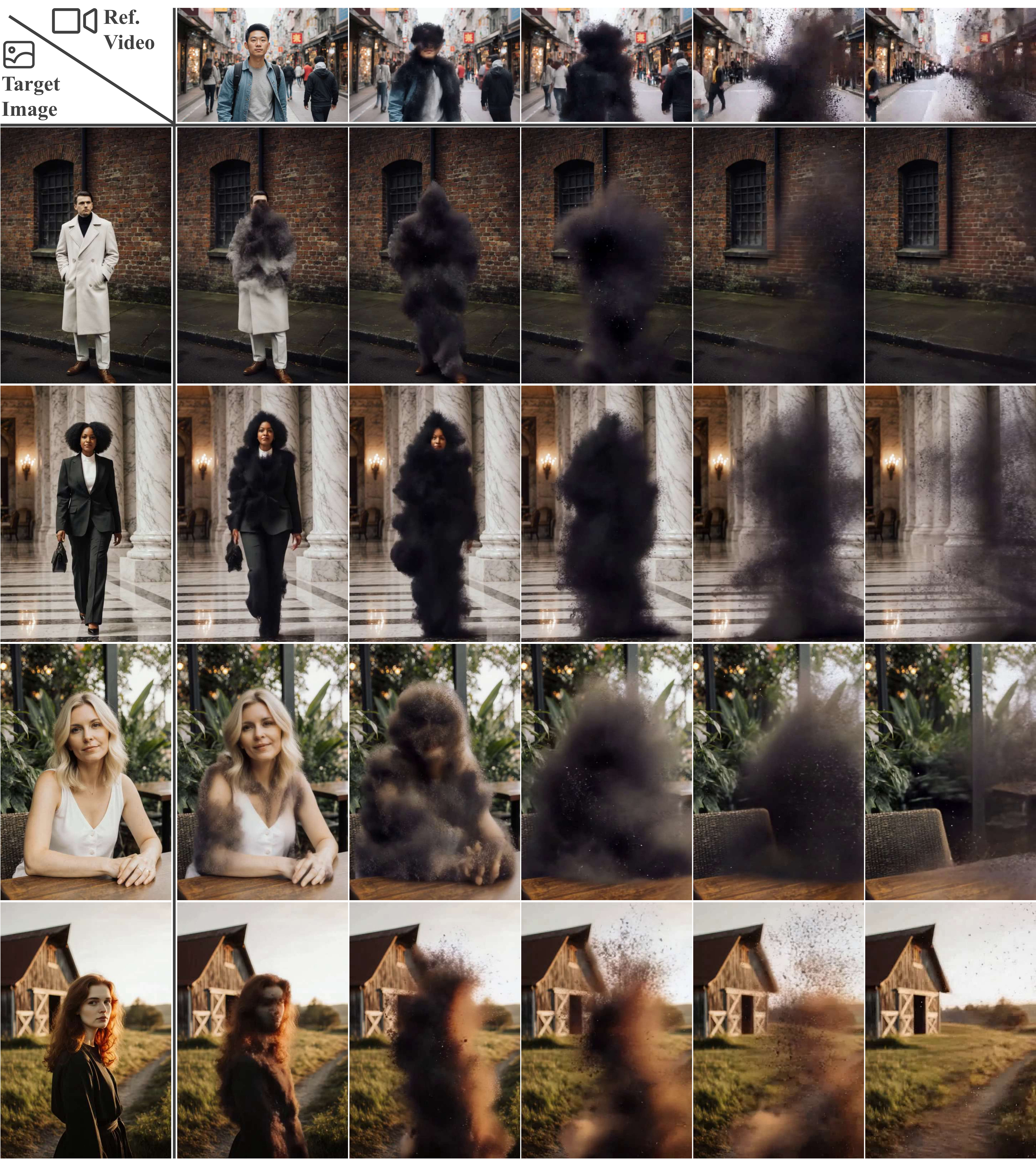}
     \vspace{-0.5cm}
    \caption{Examples of the ``Disintegration" visual effect using VFXMaster.}
    \label{fig:supp4}
\end{figure*}

\begin{figure*}
    \centering
     \includegraphics[width=1\linewidth]{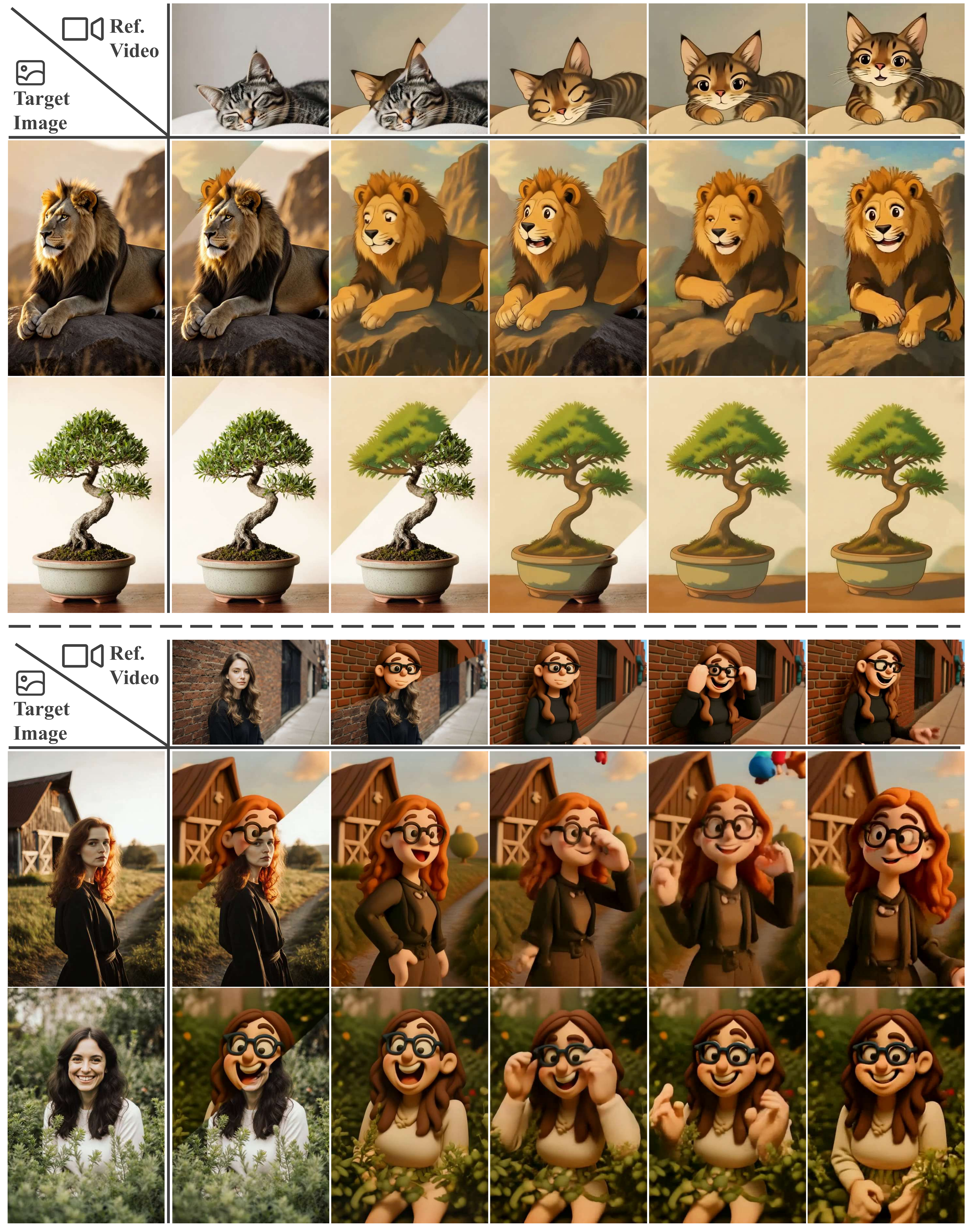}
     \vspace{-0.5cm}
    \caption{Examples of the ``Anime Couple" and ``Artistic Clay" visual effect using VFXMaster.}
    \label{fig:supp5}
\end{figure*}

\begin{figure*}
    \centering
     \includegraphics[width=1\linewidth]{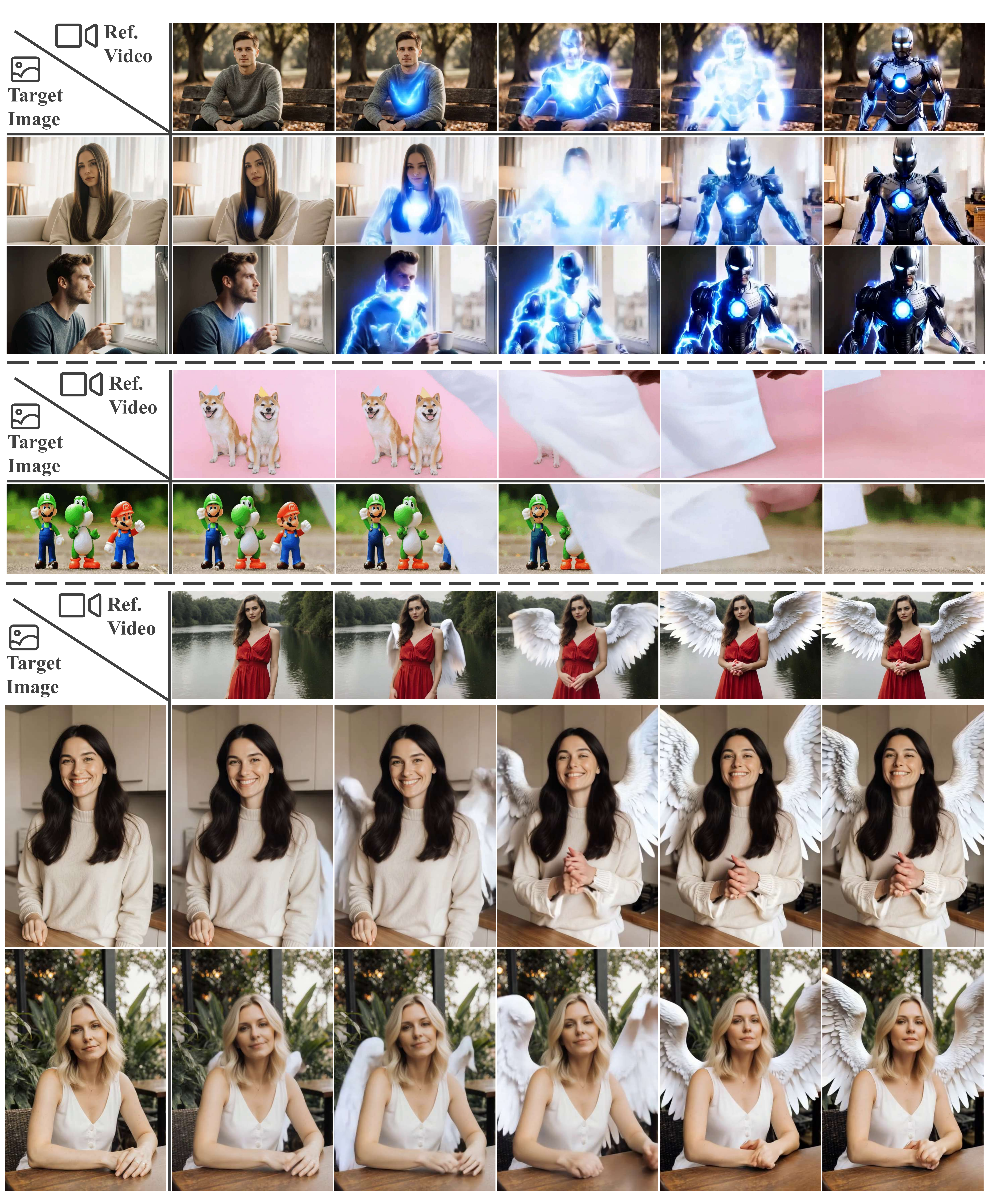}
     \vspace{-0.5cm}
    \caption{Examples of the ``The Flash", ``Tada" and ``Angle Wings" visual effect using VFXMaster.}
    \label{fig:supp6}
\end{figure*}

\begin{figure*}
    \centering
     \includegraphics[width=1\linewidth]{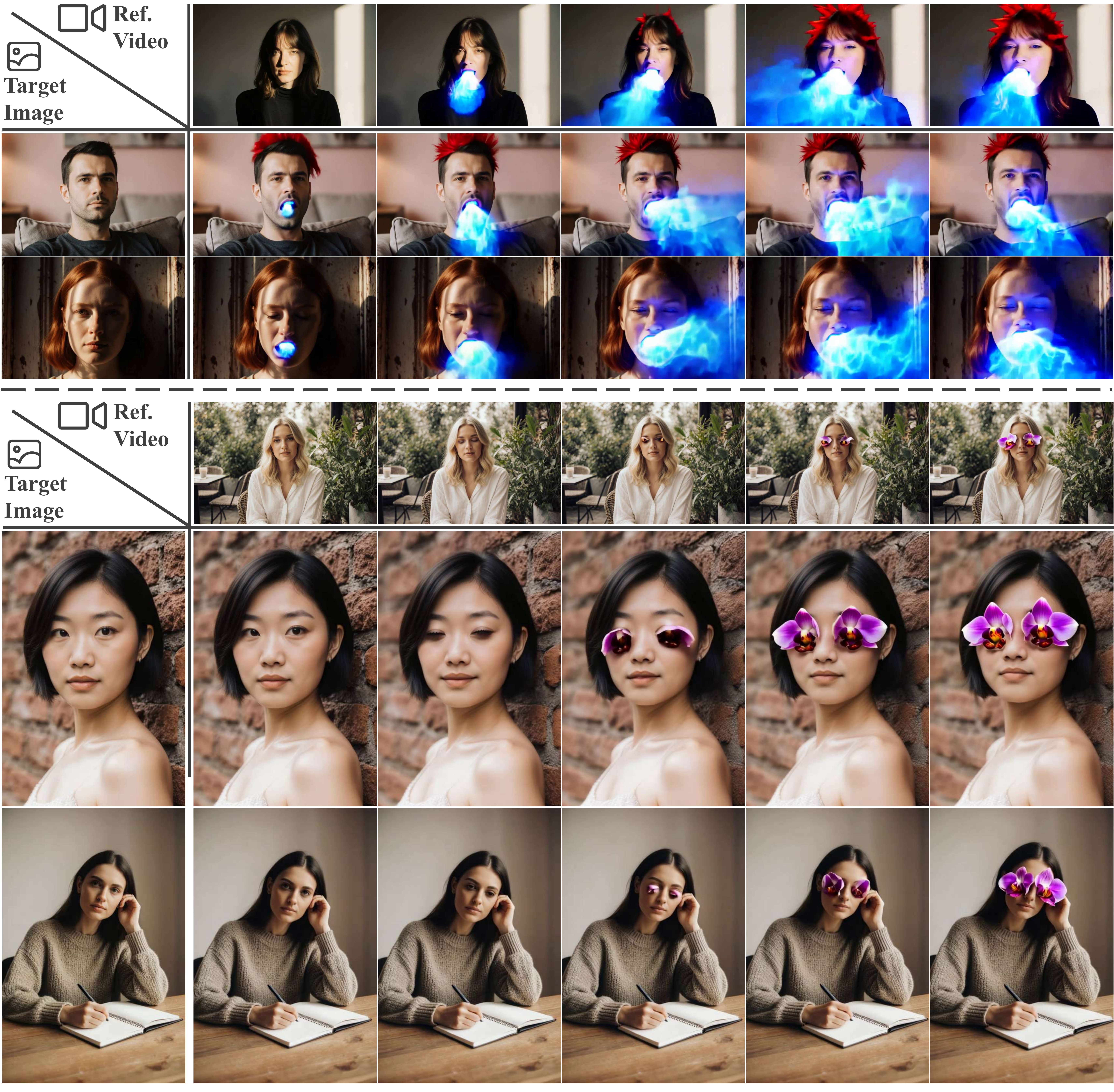}
     \vspace{-0.5cm}
    \caption{Examples of the ``Fire Breathe" and ``Floral Eyes" visual effect using VFXMaster.}
    \label{fig:supp7}
\end{figure*}

We further provide additional visual effect generation results. In-domain results are illustrated in Fig.~\ref{fig:supp1}, Fig.~\ref{fig:supp2}, Fig.~\ref{fig:supp3}, Fig.~\ref{fig:supp4} and Fig.~\ref{fig:supp5}. Out-of-domain results are illustrated in Fig.~\ref{fig:supp6}  and Fig.~\ref{fig:supp7}.

\newcolumntype{Y}{>{\centering\arraybackslash}X}
% 在泛化测试集上的更多定量对比结果
\begin{table*}[h]
\vspace{-0.4cm}
\caption{\textbf{Detailed results in Table~\ref{tab:OOD}.} Ours(one-shot) refers to the method enhanced by one-shot adaptation based on Ours.}
\vspace{0.2cm}
\centering
\tiny
\setlength\tabcolsep{0.1cm}
\begin{tabularx}{\textwidth}{@{}c|c|YYYYYYYYYYY@{}}
\toprule
\textbf{Metrics} & \textbf{Methods} & \textbf{Acid} & \textbf{Air} & \textbf{Angry\_Mode} & \textbf{Aquarium} & \textbf{Atomic} & \textbf{Balloon} & \textbf{Buddy} & \textbf{Clothes\_Rain} & \textbf{Colors\_Rain} & \textbf{Cotton} & \textbf{Fast\_Sprint} \\
\midrule
\multirow{7}{*}{\textbf{FVD$\downarrow$}} 
& Ours & 1589 & 2208 & 1753 & 2123 & \textbf{2112} & 1832 & 2454 & 1297 & 2171 &	1968 & 2554 \\
& Ours(one-shot) & \textbf{1532} & \textbf{2186} & \textbf{1657} & \textbf{1600} &	2249 & \textbf{1809} & \textbf{2445} & \textbf{1178} & \textbf{2126} & \textbf{1831} & 2496 \\
& w/o attn mask & 2534 & 3341 & 3004 & 2956 & 3460 & 2739 & 3593 & 2843 & 3060 & 4238 & 3378 \\
& w/o ref prompt & 1851 & 2409 & 2093 & 2208 & 2560 & 2192 & 2464 & 1637 & 2571 & 2258 & 2948 \\
& Ours (2k) & 2035 &3034 & 2264 & 2594 & 2992 &	2559 & 3373 & 1920 & 2633 &	3753 & 2958 \\
& Ours (4k) & 1950 & 2541 &	2101 & 2591 & 2261 & 2259 &	2909 & 1660 & 2677 & 2671 &	2495 \\
& Ours (6k) & 1702 & 2226 &	2114 & 2446 & 2211 & 1985 &	2529 & 1951 & 2451 & 2017 & \textbf{2191} \\ \midrule
\multirow{7}{*}{\textbf{\makecell[c]{Dynamic\\Degree}$\uparrow$}}
& Ours & 0.6 & 0.8 & 0.0 & 1.0 & 0.6 & 0.2 & 1.0 & 1.0 & 0.6 & 1.0 & 1.0 \\
& Ours(one-shot) & \textbf{0.6} & 0.8 & 0.4 & \textbf{1.0} & 0.6 & 0.4 & \textbf{1.0} & \textbf{1.0} & \textbf{0.6} & \textbf{1.0} & \textbf{1.0} \\
& w/o attn mask & 0.6 & \textbf{1.0} & \textbf{0.6} & 0.8 & \textbf{0.8} & \textbf{0.8} & 1.0 & 1.0 & 0.2 & 0.4 & 1.0 \\
& w/o ref prompt & 0.6 & 0.8 & 0.0 & 0.4 & 0.6 & 0.2 & 1.0 & 1.0 & 0.4 & 0.8 & 1.0 \\
& Ours (2k) & 0.4 & 0.2 & 0.0 & 0.8 & 0.6 & 0.2 & 0.6 & 1.0 & 0.2 & 0.4 & 1.0 \\
& Ours (4k) & 0.4 & 0.6 & 0.0 & 0.8 & 0.6 & 0.4 & 0.6 & 1.0 & 0.2 & 0.4 & 1.0 \\
& Ours (6k) & 0.6 & 0.8 & 0.0 & 1.0 & 0.6 & 0.2 & 0.8 &	1.0 & 0.6 & 0.8 & 1.0 \\ \midrule
\multirow{7}{*}{\textbf{EOS$\uparrow$}}
& Ours & 1.00 & 1.00 & 1.00 & 1.00 & 1.00 &	1.00 & 1.00 & 1.00 & 1.00 &	1.00 & 1.00 \\
& Ours(one-shot) & \textbf{1.00} & \textbf{1.00} & \textbf{1.00} & \textbf{1.00} & \textbf{1.00} & \textbf{1.00 }& \textbf{1.00} & \textbf{1.00} & \textbf{1.00} & \textbf{1.00} & \textbf{1.00} \\
& w/o attn mask & 0.40 & 1.00 & 1.00 & 1.00 & 1.00 & 0.80 & 0.60 & 0.60 & 0.80 & 1.00 & 1.00 \\									
& w/o ref prompt & 1.00 & 1.00 & 1.00 & 1.00 & 1.00 & 1.00 & 1.00 & 1.00 & 1.00 & 1.00 & 1.00 \\ 										
& Ours (2k) & 1.00 & 1.00 & 1.00 & 1.00 & 1.00 & 1.00 & 1.00 & 1.00 & 1.00 & 1.00 & 1.00 \\ 										
& Ours (4k) & 1.00 & 1.00 & 1.00 & 1.00 & 1.00 & 1.00 & 1.00 & 1.00 & 1.00 & 1.00 & 1.00 \\  								
& Ours (6k) & 1.00 & 1.00 & 1.00 & 1.00 & 1.00 & 1.00 & 1.00 & 1.00 & 1.00 & 1.00 & 1.00 \\ 	\midrule		
\multirow{7}{*}{\textbf{EFS$\uparrow$}}
& Ours & 0.0 & 0.6 & 0.2 & 0.6 & 0.8 & 0.6 & 0.0 & 0.6 & 0.6 & 0.8 & 0.4 \\
& Ours(one-shot) & \textbf{0.2} & \textbf{0.6} & \textbf{0.6} &	\textbf{0.8} & \textbf{1.0} &	\textbf{1.0} & \textbf{0.6} &	\textbf{1.0} & \textbf{0.8} &	\textbf{0.8} & \textbf{0.4} \\
& w/o attn mask & 0.0 & 0.2 & 0.0 & 0.0 & 0.4 & 0.2 & 0.0 & 0.0 & 0.2 & 0.2 & 0.0 \\ 											
& w/o ref prompt & 0.0 & 0.6 & 0.2 & 0.4 & 0.8 & 0.6 & 0.0 & 0.2 & 0.6 & 0.6 & 0.4 \\ 										
& Ours (2k) & 0.0 & 0.2 & 0.2 & 0.6 & 0.8 & 0.6 & 0.0 & 0.2 & 0.6 & 0.4 & 0.4 \\ 										
& Ours (4k) & 0.0 & 0.4 & 0.2 & 0.4 & 0.8 & 0.6 & 0.0 & 0.4 & 0.8 & 0.6 & 0.4 \\ 								
& Ours (6k) & 0.0 & 0.6 & 0.2 & 0.4 & 0.8 & 0.6 & 0.0 & 0.4 & 0.6 & 0.8 & 0.4 \\ 	\midrule
\multirow{7}{*}{\textbf{CLS$\uparrow$}}
& Ours & 0.8 & 1.0 & 0.8 & 0.8 & 0.8 & 0.8 & 0.8 & \textbf{0.8} & 1.0 & 1.0 & 0.4 \\
& Ours(one-shot) & \textbf{0.8} & \textbf{1.0} & \textbf{1.0} &	\textbf{1.0} & \textbf{1.0} &	\textbf{1.0} & \textbf{0.8} &	0.6 & \textbf{1.0} &	\textbf{1.0} & \textbf{0.6} \\
& w/o attn mask & 0.2 & 0.4 & 0.0 & 0.2 & 0.4 & 0.2 & 0.0 & 0.0 & 0.4 & 0.2 & 0.0 \\ 											
& w/o ref prompt & 0.8 & 1.0 & 0.6 & 0.8 & 0.8 & 0.8 & 0.8 & 0.8 & 1.0 & 0.8 & 0.4 \\ 										
& Ours (2k) & 0.8 & 1.0 & 0.8 & 0.8 & 0.8 & 0.8 & 0.8 & 0.6 & 1.0 & 1.0 & 0.4 \\ 										
& Ours (4k) & 0.8 & 1.0 & 0.8 & 0.8 & 0.8 & 0.8 & 0.8 & 0.8 & 1.0 & 1.0 & 0.4 \\ 								
& Ours (6k) & 0.8 & 1.0 & 0.8 & 0.8 & 0.8 & 0.8 & 0.8 & 0.8 & 1.0 & 1.0 & 0.4 \\ 	\midrule
\textbf{Metrics} & \textbf{Methods} & \textbf{Hair} & \textbf{Flight} & \textbf{Illustration} & \textbf{BOOM} & \textbf{Mask} & \textbf{Pizza} & \textbf{Shadow} & \textbf{Spirit\_Animal} & \textbf{To\_Monkey} & \textbf{Avg.} \\ \midrule

\multirow{7}{*}{\textbf{FVD$\downarrow$}} 
& Ours & \textbf{2449} &	2960 & 1588 & 2442 & 3101 &	1898 & 1927 & 2664 & 1963 &	2153 \\
& Ours(one-shot) & 2602 & \textbf{2384} &	\textbf{1330} & \textbf{2366} & \textbf{3003} & \textbf{1841} & \textbf{1895} & \textbf{2513} & \textbf{1889} & \textbf{2047} \\
& w/o attn mask & 4554 & 4158 &	3140 & 3754 & 4650 & 2967 & 3123 & 3601 & 4242 & 3467 \\
& w/o ref prompt & 3571 & 3163 & 1921 &	3047 &3498 & 2266 &	2214 & 2664 & 2132 & 2483 \\
& Ours (2k) & 3837 & 3730 &	2374 & 3457 & 4521 & 2496 & 2379 & 3407 & 2440 & 2938 \\
& Ours (4k) & 3859 & 2860 &	1904 & 3031 & 4368 & 2173 &	2068 & 2935 & 2126 & 2572 \\
& Ours (6k) & 2528 & 2935 & 1872 & 3081 & 3736 & 2171 & 2011 & 2807 & 2037 & 2350 \\ \midrule
\multirow{7}{*}{\textbf{\makecell[c]{Dynamic\\Degree}$\uparrow$}}
& Ours & 1.0 & 1.0 & 0.6 & 1.0 & 1.0 & 1.0 & 0.4 & 1.0 & 1.0 & 0.79 \\
& Ours(one-shot) & \textbf{1.0} & \textbf{1.0} & \textbf{0.6} & \textbf{1.0} & \textbf{1.0} & \textbf{1.0} & \textbf{0.8} &	\textbf{1.0} & \textbf{1.0} &	\textbf{0.84} \\
& w/o attn mask & 1.0 & 1.0 & 0.6 &	1.0 & 1.0 & 1.0 & 0.4 & 1.0 & 1.0 &	0.81 \\
& w/o ref prompt & 1.0 & 1.0 & 0.6 & 1.0 & 1.0 & 1.0 & 0.4 & 1.0 & 1.0 & 0.74 \\
& Ours (2k) & 0.8 & 1.0 & 0.2 & 0.4 & 1.0 & 0.8 & 0.4 & 1.0 & 1.0 & 0.60 \\
& Ours (4k) & 0.8 & 1.0 & 0.4 &	0.4 & 1.0 & 0.8 & 0.4 & 1.0 & 1.0 & 0.64 \\
& Ours (6k) &0.8 & 1.0 & 0.4 & 0.6 & 1.0 & 1.0 & 0.4 & 1.0 & 1.0 & 0.70 \\ \midrule
\multirow{7}{*}{\textbf{EOS$\uparrow$}}
& Ours & 1.00 & 1.00 & 1.00 & 1.00 & 1.00 &	1.00 & 1.00 & 1.00 & 1.00 & 1.00 \\
& Ours(one-shot) & \textbf{1.00} & \textbf{1.00} & \textbf{1.00} & \textbf{1.00} & \textbf{1.00} & \textbf{1.00} & \textbf{1.00} & \textbf{1.00} & \textbf{1.00} & \textbf{1.00} \\
& w/o attn mask & 0.80 & 1.00 & 1.00 & 1.00 & 1.00 & 1.00 & 0.80 & 1.00 & 1.00 & 0.89 \\ 									
& w/o ref prompt & 1.00 & 1.00 & 1.00 & 1.00 & 1.00 & 1.00 & 1.00 & 1.00 & 1.00 & 1.00 \\ 										
& Ours (2k) & 1.00 & 1.00 & 1.00 & 1.00 & 1.00 & 1.00 & 0.80 & 1.00 & 1.00 & 0.97 \\ 										
& Ours (4k) & 1.00 & 1.00 & 1.00 & 1.00 & 1.00 & 1.00 & 1.00 & 1.00 & 1.00 & 0.99 \\  								
& Ours (6k) & 1.00 & 1.00 & 1.00 & 1.00 & 1.00 & 1.00 & 1.00 & 1.00 & 1.00 & 1.00 \\ 	\midrule
\multirow{7}{*}{\textbf{EFS$\uparrow$}}
& Ours & 0.8 & 0.6 & 0.0 & 0.2 & 0.0 & 1.0 & 0.4 & 0.8 & 0.4 & 0.47 \\
& Ours(one-shot) & \textbf{1.0} & \textbf{0.6} & \textbf{0.6} & \textbf{0.4} & \textbf{0.2} & \textbf{1.0} & \textbf{1.0} & \textbf{0.8} & \textbf{0.6} & \textbf{0.70} \\
& w/o attn mask & 0.0 & 0.0 & 0.0 & 0.2 & 0.0 & 0.2 & 0.0 & 0.4 & 0.0 & 0.11 \\ 									
& w/o ref prompt & 0.8 & 0.6 & 0.0 & 0.0 & 0.0 & 1.0 & 0.2 & 0.8 & 0.2 & 0.40 \\ 										
& Ours (2k) & 0.4 & 0.4 & 0.0 & 0.0 & 0.0 & 0.8 & 0.4 & 0.6 & 0.2 & 0.34 \\ 										
& Ours (4k) & 0.8 & 0.4 & 0.2 & 0.0 & 0.0 & 0.8 & 0.4 & 0.6 & 0.2 & 0.40 \\  								
& Ours (6k) & 0.8 & 0.6 & 0.0 & 0.0 & 0.0 & 0.6 & 0.6 & 0.8 & 0.2 & 0.42 \\ 	\midrule
\multirow{7}{*}{\textbf{CLS$\uparrow$}}
& Ours & 0.6 & 1.0 & \textbf{1.0} & 0.4 & 0.4 & 1.0 & 1.0 & 0.8 & 0.6 & 0.79 \\
& Ours(one-shot) & \textbf{0.8} & \textbf{1.0} & 0.8 & \textbf{0.6} & \textbf{0.4} & \textbf{1.0} & \textbf{1.0} & \textbf{1.0} & \textbf{1.0} & \textbf{0.87} \\
& w/o attn mask & 0.2 & 0.6 & 0.4 & 0.0 & 0.0 & 0.8 & 0.4 & 0.0 & 0.4 & 0.24 \\ 									
& w/o ref prompt & 0.6 & 1.0 & 0.8 & 0.4 & 0.4 & 1.0 & 0.8 & 1.0 & 0.6 & 0.76 \\ 										
& Ours (2k) & 0.6 & 1.0 & 1.0 & 0.4 & 0.4 & 1.0 & 1.0 & 0.6 & 0.6 & 0.77 \\ 										
& Ours (4k) & 0.6 & 1.0 & 0.6 & 0.4 & 0.4 & 1.0 & 1.0 & 0.8 & 0.4 & 0.76 \\  								
& Ours (6k) & 0.6 & 1.0 & 0.8 & 0.4 & 0.4 & 1.0 & 1.0 & 1.0 & 0.6 & 0.79 \\
\bottomrule

\end{tabularx}
\label{tab:generalized_detail}
\vspace{-0.5cm}
\end{table*}